\documentclass{article}

\usepackage[utf8]{inputenc} 
\usepackage[T1]{fontenc}    
\usepackage{hyperref}       
\usepackage{url}            
\usepackage{booktabs}       
\usepackage{amsfonts}       
\usepackage{nicefrac}       
\usepackage{microtype}      
\usepackage[dvipsnames,table]{xcolor}

\usepackage[accepted]{icml2026}

\usepackage{amsmath,amssymb}
\usepackage{mathtools}
\usepackage{siunitx}
\usepackage[capitalize,noabbrev,nameinlink]{cleveref}
\usepackage{amsthm}
\usepackage{multirow}
\usepackage[pdftex]{graphicx}
\usepackage{subcaption}
\usepackage{bm}
\usepackage{wrapfig}
\usepackage{float}
\usepackage{array}
\usepackage{stfloats}
\usepackage{enumitem}
\setlist[itemize]{leftmargin=1em}
\usepackage{caption}
\crefname{section}{section}{sections}
\crefname{figure}{figure}{figures}
\crefname{table}{table}{tables}
\crefname{appendix}{appendix}{appendices}
\crefname{equation}{equation}{equations}

\usepackage{algorithm}
\usepackage{algpseudocode}
\algrenewcommand\algorithmicrequire{\textbf{Input:}}
\algrenewcommand\algorithmicensure{\textbf{Output:}}

\icmltitlerunning{Byte Pair Encoding for Efficient Time Series Forecasting}

\begin{document}

\twocolumn[
  \icmltitle{Byte Pair Encoding for Efficient Time Series Forecasting}

\begin{icmlauthorlist}
\icmlauthor{Leon G\"otz}{vw,tum}
\icmlauthor{Marcel Kollovieh}{tum}
\icmlauthor{Stephan G\"unnemann}{tum}
\icmlauthor{Leo Schwinn}{tum,hel}
\end{icmlauthorlist}

\icmlaffiliation{vw}{Volkswagen AG}
\icmlaffiliation{tum}{Technical University of Munich}
\icmlaffiliation{hel}{Helmholtz AI}
\icmlcorrespondingauthor{Leon G\"otz}{leon.goetz@volkswagen.de}

\icmlkeywords{Time Series, Tokenization, Transformer, Foundation Model}

  \vskip 0.3in
]

\printAffiliationsAndNotice{}

\begin{abstract}
Existing time series tokenization methods predominantly encode a constant number of samples into individual tokens. This inflexible approach can generate 
excessive tokens for even simple patterns like extended constant values, resulting in  
substantial computational overhead. Inspired by the success of byte pair encoding, we propose the first pattern-centric tokenization scheme for time series analysis. Based on a discrete vocabulary of frequent motifs, our method merges samples with underlying patterns into tokens, compressing time series adaptively. Exploiting our finite set of motifs and the continuous properties of time series, we further introduce conditional decoding as a lightweight yet powerful post-hoc optimization method, which requires no gradient computation and adds no computational overhead. On recent time series foundation models, our motif-based tokenization improves forecasting performance by \SI{40}{\percent} and boosts efficiency by \SI{2314}{\percent} on average. Conditional decoding further reduces MSE by up to \SI{48}{\percent}. In an extensive analysis, we demonstrate the adaptiveness of our tokenization to diverse temporal patterns, its generalization to unseen data, and its meaningful token representations capturing distinct time series properties, including statistical moments and trends.
\end{abstract}

\vspace{-1\baselineskip}
\section{Introduction}
\vspace{-0.3\baselineskip}
\begin{figure}
    \centering
    \includegraphics[width=6.71cm,trim={0cm 0cm 0cm 0cm},clip]{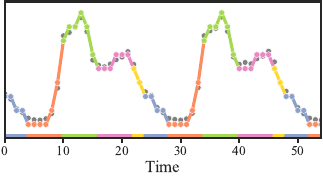}
    \vspace{-0.4\baselineskip}
    \caption{
    Motif-based tokenization transforms time series data (gray) through a two-step process: 1) quantizing samples into discrete bins, 2) merging recurring patterns of variable length into representative motifs (identical motifs share the same color). Motif repetition is highlighted by their x‑axis projection.}
    \label{fig:tokenized_sequence_zoomed}
    \vspace{-1.3\baselineskip}
\end{figure}

Transformer architectures have gained increasing relevance in time series processing, demonstrating impressive performance. Here, a key prerequisite for strong performance is effective tokenization -- dividing the input into smaller units and embedding them in a high-dimensional space.

Yet, current tokenization schemes in time series processing exhibit considerable limitations: Early works embed each individual time step as a token, creating a fundamentally inefficient representation, where every token captures little temporal information. This results in very long token sequences, imposing a substantial computational burden in the transformer architecture \citep{goetz2025localmerging}. Splitting the time series into fixed-length subsequences, called patches, mitigates both issues \citep{nie2023patchtst}. However, rigid patches cannot adapt to diverse temporal patterns of varying lengths and complexities \citep{woo2024unifiedtraining}.

Inspired by adaptive pattern-based tokenization schemes in natural language processing (NLP) \citep{sennrich2016bpetokenization}
, we go beyond previous work and propose the first pattern-centric tokenization for time series, illustrated in \cref{fig:tokenized_sequence_zoomed}. Our contribution is threefold:

\vspace{2pt}
\textbf{Adaptive tokenization for time series}\hspace{1.8mm} We provide a novel tokenization strategy based on a discrete vocabulary of frequent time series motifs. Our method merges samples with underlying patterns into single tokens, enabling adaptive compression while maintaining a small upper-bounded discretization error. On the recently proposed Chronos foundation model, our tokenization improves forecasting performance by \SI{40}{\percent} and boosts efficiency by \SI{2314}{\percent} on average in a zero-shot setting.

\vspace{2pt}
\textbf{Conditional decoding}\hspace{1.8mm} We introduce conditional decoding as a post-hoc optimization method to further improve forecasting performance by exploiting the continuous properties of time series to effectively remove the discretization error induced by our motifs.  Conditional decoding is lightweight, requires no gradient computation, introduces no additional overhead during inference, and can be combined with any pretrained time series model with a discrete output vocabulary. We demonstrate its effectiveness in large foundation models, increasing forecasting performance up to \SI{48}{\percent}. 

\vspace{2pt}
\textbf{Empirical analysis}\hspace{1.8mm} In an extensive empirical study, we demonstrate the zero-shot generalization capability of our tokenizer and its ability to automatically adapt to diverse temporal patterns and datasets. We link distinct time series characteristics, including statistical moments and trends, to our token representations and show that complex motifs benefit forecasting quality.

\vspace{0.25\baselineskip}
\section{Related work}
\vspace{0.1\baselineskip}
In recent years, transformer models have shown impressive performance in time series forecasting. While initial work focuses on efficient attention mechanisms and domain-specific architectures \citep{wu2021autoformer, zhou2022fedformer}, universal foundation models have been proposed lately \citep{garza2023timegpt1, das2024timesfm, rasul2024lagllama,chronos, woo2024unifiedtraining, goswami2024moment, gao2024units,cohen2024toto,liu2024moiraimoe,liu2024timer,liu2025sundial}. These models are usually trained on billions of tokens and exhibit high zero-shot performance. However, all these transformer architectures rely on two basic tokenization techniques: using every sample as a token or extracting fixed-length patches from time series.

\textbf{Sample-based tokens}\hspace{1.8mm} Most early works on transformer models for time series processing extract tokens for every time step, usually as a slice of a multivariate time series \citep{informer, wu2021autoformer, zhou2022fedformer, liu2022pyraformer,liu2022non, cirstea2022triformer}. These tokens are linearly transformed into a continuous embedding space. Inspired by the success of discrete token embeddings in NLP, the recently proposed Chronos foundation model \citep{chronos} quantizes a univariate time series into bins and embeds them using learned vectors. This way, the authors transform forecasting from a regression task to classifying the next time step from a discrete vocabulary \citep{torgo1997regressiontoclassification}. \citet{masserano2024wavelettok} use a wavelet-transformation-based approach for tokenization.
Generating tokens for every time step has two major limitations: First, the large number of tokens imposes a substantial computational burden in transformers, especially for long sequence processing \citep{godahewa2021monash,chronos}. Second, every token captures only little information about temporal patterns \citep{chen2025acloserlook}.

\textbf{Patch-based tokens}\hspace{1.8mm} Inspired by the success of patching in computer vision \citep{dosovitskiy2021vit}, \citet{nie2023patchtst} adapt this approach to time series, where multiple samples of a univariate time series are combined into individual tokens. 
Most subsequent works embed the patches into a continuous space using learned transformations \citep{zhang2023crossformer, nie2023patchtst, wang2024timexer, wu2024perimidformer, das2024timesfm, woo2024unifiedtraining, goswami2024moment, gao2024units, cohen2024toto,auer2025tirex,liu2024moiraimoe,liu2024timer,liu2025sundial}. More advanced approaches learn a discrete codebook of patches \citep{talukder2024totem, chen2024sdformer} using vector quantized variational autoencoder approaches \citep{oord2017vqvae}. 
Patches generally compress the time series and capture local temporal information.
However, due to their fixed length and stride, rigid patches can not adapt to varying temporal patterns in a sequence. This is particularly important for foundation models, as they aim to generalize to previously unseen data in zero-shot settings. To mitigate this, \citet{woo2024unifiedtraining} utilize different patch lengths for datasets sampled in different granularities, e.g., minutely or hourly. Their approach requires training of a new embedding transformation for every granularity. \citet{shi2025timemoe} use multiple predefined patch lengths to predict time series in a mixture of experts approach and \citet{wang2025lightgts} derive the patch length from the single dominant frequency in a time series. Even these approaches fail to capture diverse patterns of arbitrary lengths within a sequence (see \cref{sec:adaptive_compression}).

\textbf{Motif-based tokens}\hspace{1.8mm} Motif-based tokenization utilizes a discrete vocabulary of recurring patterns. 
In NLP, byte pair encoding hierarchically extracts pairs of character-bytes to tokenize a sentence \citep{shibata1999bpetextcompression,sennrich2016bpetokenization}. \citet{elsner2024multidimensionalbytepairencoding} extend this concept from 1d-sequences to tokenizing images. 
Moreover, tokenization based on discrete motifs has proven to be a good inductive bias for high-dimensional distribution learning as it reduces the combinatorial complexity \citep{sommer2023powerofmotifs}. Similarly, classical time series literature explored symbolization and pattern discovery techniques \citep{lin2003sax,berndt1994timeseries_dtw}. 
Yet, data-dependent tokenization techniques as proposed in this work remain unexplored for machine-learning-based time series analysis.

\begin{figure*}[h]
    \centering
    \includegraphics[width=\textwidth,trim={0in 5.93in 0.1in 0in},clip]{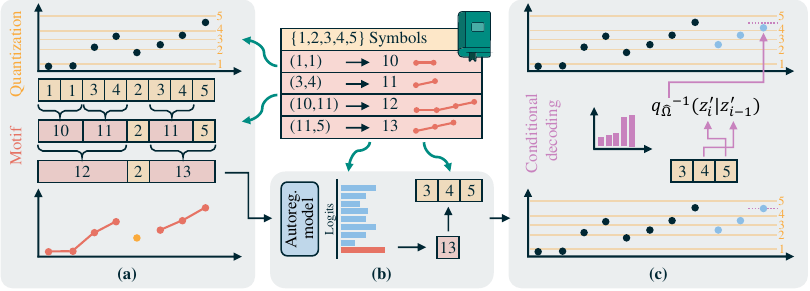}
    \caption{\textbf{(a)} Our motif-based tokenization first quantizes a time series into symbols and finds recurring motifs as tokens, building a discrete vocabulary. \textbf{(b)} Based on the compressed motif sequence, a neural network forecasts the time series through a categorical distribution over our vocabulary. \textbf{(c)} Finally, we propose conditional decoding to reduce the discretization error when transforming tokens back to their continuous representation.}
    \vspace{-0.5\baselineskip}
    \label{fig:method}
\end{figure*}

\section{An adaptive tokenization approach for \mbox{time series}}
\label{sec:method_motif_tok}
Despite recent advances in time series processing, current tokenization methods lack efficiency or fail to capture distinct temporal patterns within sequences \citep{ekambaram2024tinytimemixers, woo2024unifiedtraining}. We propose an efficient tokenization method using a vocabulary of frequent motifs as depicted in \cref{fig:method}. Our algorithm combines samples with underlying patterns of varying complexity into single tokens. 
Its adaptive compression of time series enables efficient long sequence processing. We list pseudocode in \cref{appendix:algorithms}.

Let $\mathcal{D} = \{z^i\}_{i=1}^N$ be a family indexed by $i\!=\!1, \dots, N$ of $N$ univariate real-valued time series $z = (z_1, \dots , z_n) \in\mathbb{R}^{n}$ of length $n$. We normalize each series to have zero mean and unit standard deviation. A neural network $\mathbf{f}_{\boldsymbol{\theta}}:\mathbb{N}^{t_{\mathrm{in}}}\rightarrow \mathbb{N}^{t_{\mathrm{out}}}$ with parameters $\boldsymbol{\theta}$ predicts $t_{\mathrm{out}}$ token IDs from $t_{\mathrm{in}}$ token IDs. Thereby, the tokens are generated by our tokenizer $\mathbf{g}:\mathbb{R}^{n}\rightarrow \mathbb{N}^{t}$ from a time series $z$. 

Our tokenization consists of two steps:
\begin{equation}
\mathbf{g}(z) = \mathbf{m}_\Psi \circ \mathbf{q}_\Omega(z)
\end{equation}
\vspace{-0.7\baselineskip}
where:
\begin{align*}
\begin{array}{@{}l@{\quad}m{0.6\linewidth}@{}}
\mathbf{q}_\Omega:\mathbb{R}^{n}\rightarrow \mathbb{N}^{n}  \quad  & \small\textnormal{quantizes the time series into a sequence of discrete symbols}, \\
\mathbf{m}_\Psi:\mathbb{N}^{n}\rightarrow \mathbb{N}^{t} \quad & \small\textnormal{compresses the sequence based on a discrete vocabulary of temporal motifs}, \\
\Omega, \Psi \quad & \small\textnormal{vocabulary of quantized symbols and motifs, respectively}.
\end{array}
\end{align*}

\subsection{Discretization of real-valued time series}
\label{sec_method:quantization}
Generalizing the approach from \citet{chronos}, we sample $M$ equiprobable discretization intervals \mbox{$\Omega = \{C^{-1} \left(\frac{j}{M}\right)\}_{j=1}^M$}, where $C^{-1}$ is the inverse cumulative distribution of the probability distribution $P$. In practice, we experiment with truncated uniform distributions in $[\omega_{\mathrm{lb}}, \omega_{\mathrm{ub}}]$, Gaussian distributions, and the precise data distribution $P(\mathcal{D})$ for binning. Utilizing the boundaries, we encode the time series $z$ into a sequence of discrete symbols: 

\vspace{-1\baselineskip}
\begin{equation}
\begin{aligned}
\mathbf{q}_\Omega(z) &= \{ q_\Omega(z_i) \: |\: z_i \in z \} \text{,}\\[6pt]
\text{\!where}\quad 
q_\Omega(z_i) &= \begin{cases} 
1 & \text{if } z_i \leq \omega_1 \\
j & \text{if } \omega_{j-1} < z_i \leq \omega_{j}
\end{cases}
\text{\!, \:}\omega_{j} \in \Omega\:.
\end{aligned}
\end{equation}

For decoding symbol IDs back to time series samples, we use $\hat{\Omega} = \{C^{-1} \left(\frac{j-0.5}{M}\right)\}_{j=1}^M$, where $\hat{\omega}_j \in \hat{\Omega}$ is the probabilistic center of $[\omega_{j-1}, \omega_j]$.
Within the tokenization range, the quantization error can be upper bounded as $\delta_{\mathrm{max}} = \max_{1 < j \leq M} \max(\hat{\omega}_j - \omega_{j-1},\: \omega_{j} - \hat{\omega}_j)$. For uniform binning, the probabilistic center is equal to the geometric center, and the maximum error simplifies to $\delta_{\mathrm{max}} = (\omega_{\mathrm{ub}} - \omega_{\mathrm{lb}})(2M)^{-1}$. Besides the $M$ token IDs representing quantized time series samples, we further introduce a masking token \texttt{MASK} to account for missing samples and an \texttt{EOS} token, which we insert at the end of time series.

\subsection{Vocabulary of temporal motifs}
Originally proposed for compressing raw byte sequences \citep{gage1994bpeoriginal}, byte pair encoding has been widely used in NLP to compress character sequences into subwords \citep{sennrich2016bpetokenization}. Here, we generalize the byte pair compression algorithm to extract temporal patterns from our discretized time series.
To this end, we iteratively build a vocabulary $\Psi$ of frequent time series motifs: Given a dataset $\mathcal{D^\prime} = \{\mathbf{q}_\Omega(z^i) \: | \: z^i \in \mathcal{D}\}$ of quantized time series, we extract the most frequent adjacent token IDs $(z^\prime_i, z^\prime_{i+1})$, assigning a new token ID $z^\prime_{\mathrm{new}}$, which we add to our set of patterns $\Psi$:

\vspace{-0.5\baselineskip}
\begin{equation}
\label{eq:motif_vocabulary}
\Psi^{(l+1)} \leftarrow \Psi^{(l)} \cup \{(z^\prime_i, z^\prime_{i+1}) \rightarrow z^\prime_{\mathrm{new}}\}\:.
\end{equation}

This process hierarchically finds distinct temporal motifs as discrete tokens and is locally optimal in every step. We build our vocabulary until the new tokens occur less frequently than $p_{\mathrm{min}}$ in $\mathcal{D^\prime}$. This ensures that a minimum number of occurrences are available for a neural network to learn the motifs. Leveraging our vocabulary, we compress a quantized time series into a sequence of motifs: 

\vspace{-1\baselineskip}
\begin{equation}
\label{eq:tokenization_m}
\mathbf{m}_\Psi(z^\prime) = \{ \psi(z^\prime) \: |\: \psi \in \Psi \}, \quad \mathbf{m}_\Psi:\mathbb{N}^{n}\rightarrow \mathbb{N}^{t}\:.
\end{equation}
\vspace{-1\baselineskip}

The compression is highly flexible as motifs of different lengths and complexities are mapped to single tokens. We define the average compression at the sequence level as $\bar{c}=n/t$. Our algorithm has linear complexity in sequence length $O(|\Psi|\cdot n)$, enabling long sequence processing.

\vspace{-0.15\baselineskip}
\subsection{Conditional decoding}
\label{sec:method_cond_decode}
\vspace{-0.15\baselineskip}
We propose our novel conditional decoding to universally improve the forecasting quality of models with discrete output vocabularies. To decode a token sequence, such as the predictions of a model, we invert the tokenization 
$\mathbf{g}$. In this process when inverting $\mathbf{q}_\Omega$, we previously leveraged the bin centers 
$\hat{\omega}_j \in \hat{\Omega}$ to transform a quantized sequence $z^\prime$ back to a time series  
$\hat{z}$. We introduce conditional decoding to reduce the overall quantization error. Specifically, 
we decode quantized time series samples $z^\prime_i$ conditioned on the previous sample \smash{$\hat{z}_i = q^{-1}_{\hat{\Omega}} (z^\prime_i \: | \: z^\prime_{i-1})$}. 
To this end, we set parameters \smash{$\hat{\Omega}= \{ \hat{\omega}_{j,k} \mid j, k \in \{1, \dots, M\} \}$}, where $q^{-1}_{\hat{\Omega}} (z^\prime_i=j \: | \: z^\prime_{i-1}=k)=\hat{\omega}_{j,k}$ to minimize $\lVert z_i - \hat{z}_i\rVert^2_2$:

\vspace{-0.6\baselineskip}
\begin{equation}
\begin{aligned}
\min_{\hat{\Omega}} \sum_{(z, z^\prime) \in \mathbf{D}} \sum_{i=2}^n \big\lVert z_i - q^{-1}_{\hat{\Omega}}(z^\prime_i \: | \: z^\prime_{i-1})\big\rVert^2_2\text{,}\\[6pt]
\text{where}\quad  \mathbf{D} = \{ (\mathcal{D}_i, \mathcal{D}^\prime_i) \: | \: i \in \{1, \dots , N \}\}
\end{aligned}
\end{equation}
\vspace{-0.6\baselineskip}

consists of corresponding real-valued and quantized time series. Thereby, a single parameter $\hat{\omega}_{j,k}$ is given by the mean of the underlying time series samples $\tilde{z}$, minimizing the squared error:

\vspace{-1.6\baselineskip}
\begin{equation}
\label{eq:cond_decode_solution}
\begin{aligned}
\hat{\omega}_{j,k} &= \frac{1}{|\mathcal{D}_{j,k}|} \sum_{\tilde{z} \in \mathcal{D}_{j,k}} \tilde{z}\text{,}\\[6pt]
\text{where}\quad 
\mathcal{D}_{j,k} &= 
\{z_i \: | \: (z, z^\prime) \in \mathbf{D}, z^\prime_i=j, z^\prime_{i-1}=k\} \:.
\end{aligned}
\end{equation}
\vspace{-1.0\baselineskip}

Intuitively, we adopt a unigram model to exploit the unique properties of our tokenization: the finite set of discrete symbols and the underlying continuous time series samples. 
Conditional decoding is lightweight and requires no gradient computation as we solve analytically for the global optimum. Further, it adds no additional inference cost\footnote{It uses the same $O(1)$ dictionary lookup as normal decoding.} and is very small in practice with only $M^2$ parameters $\hat{\omega}_{j,k} \in \hat{\Omega}$ relying on the first-order Markov assumption\footnote{We analyze higher-order models in \cref{appendix:conditional_decoding}.}. Conditional decoding can be combined with any pretrained time series model with a discrete output vocabulary and considerably improves forecasting performance in our experiments.

\vspace{-0.15\baselineskip}
\subsection{Model architecture}
\vspace{-0.15\baselineskip}
\looseness=-1
As we represent continuous time series as a sequence of discrete motifs, we can rely on recent advances in transformer architectures in natural language processing. These architectures transform token IDs from our discrete vocabulary $\mathcal{V} = \Omega \cup \Psi \cup \{\texttt{MASK}, \texttt{EOS}\}$ into $d$-dimensional space using learned embedding tables $E \in \mathbb{R}^{|\mathcal{V}| \times d}$. 
We optimize the parameters $\boldsymbol{\theta} \in \boldsymbol{\Theta}$ of our model $\mathbf{f}_{\boldsymbol{\theta}}$ on autoregressive next token prediction of our tokenized sequence \mbox{$z^{\prime\prime} = \mathbf{g}(z)$.} Our model thereby predicts a categorical distribution $p(z^{\prime\prime}_{t_{\mathrm{in}}+1} \: |\: z^{\prime\prime}_{1:t_{\mathrm{in}}})$ over our finite vocabulary of time series motifs $\mathcal{V}$. We impose a cross-entropy loss for distribution learning. 
To this end, we transform the regression task to a classification \citep{torgo1997regressiontoclassification}. The discrete set of possible motifs reduces the combinatorial complexity and has proven to be a good inductive bias for distribution learning in the bio-medical domain \citep{sommer2023powerofmotifs}. In contrast to prior work \citep{chronos}, our tokenizer enhances the efficiency as both model input and generated tokens are compressed time series representations. Models can utilize longer contexts while requiring fewer autoregressive iterations for a given prediction horizon. This is especially important for large foundation models and long sequence processing, imposing substantial computational demands.

\vspace{-0.15\baselineskip}
\section{Experiments}
\label{sec:experiments}
\vspace{-0.15\baselineskip}
We systematically train different tokenizers and foundation models and evaluate them on \num{7} time series datasets in a zero-shot setting, demonstrating advantages of our motif-based representation over tokenizing every sample or utilizing patches. In \cref{appendix:experiments}, we provide further details.

\textbf{Datasets}\hspace{1.8mm} For training our models and tokenizers, we utilize the recently proposed Chronos dataset \citep{chronos}. It contains \num{11}\,M time series with over \num{11}\,B samples. Due to its diverse nature and size, this dataset is well-suited for training foundation models. 
We base our zero-shot evaluation on \num{7} commonly used time series datasets: ETTh1, ETTm1, Weather, Electricity, Traffic, Solar, and Fev-bench \citep{godahewa2021monash,shchur2026fevbench}. Note that Fev-bench is a combination of multiple datasets. We specify the exact configuration in~\cref{appendix:experiments}.

\textbf{Tokenizers}\hspace{1.8mm} We leverage 3 tokenizers with different numbers of quantization bins $M$. Further, we utilize a truncated uniform distribution from $\omega_{\mathrm{lb}} = -5$ to $\omega_{\mathrm{ub}}=5$ for binning, spanning a range of \num{5} standard deviations.  
As a result, our tokenizers in \cref{tab:tokenizers} feature different compression ratios, vocabulary sizes, and discretization errors. We build their vocabulary $\Psi$ on the same \num{100000} randomly selected time series from the Chronos dataset with a total of \num{100}\,M samples. To allow the model to learn all tokens, we constrain the motifs to occur at least $p_{\mathrm{min}}=1000$ times in the compressed data. In \cref{sec:vocab_complexity,sec:dataset_size,appendix:preprocessing_strategies}, we systematically ablate these choices.

\vspace{-0.2\baselineskip}
   \begin{table}[h]
      \caption{Tokenizers on the Chronos dataset with different quantization bins, vocabulary size, discretization error, and compression.}
      \label{tab:tokenizers}
      \centering
      \vspace{-0.5\baselineskip}
        \resizebox{6cm}{!}{
        \begin{tabular}{lrrrr}
            \toprule 
            Compression & $M$ & \multicolumn{1}{c}{$|\mathcal{V}|$} & $\delta_{\mathrm{max}}$ & \multicolumn{1}{c}{$\bar{c}$}\\ 
            \midrule 
            low & \num{126} & \num{2445} & \num{0.040} & \num{2.08}\\
            medium & \num{37} & \num{1675} & \num{0.135} & \num{3.18}\\
            high & \num{22} & \num{1373} & \num{0.227} & \num{4.06}\\
            \bottomrule 
        \end{tabular}
        }
        \vspace{-0.5\baselineskip}
    \end{table}

\textbf{Models}\hspace{1.8mm} In our experiments, we explore our tokenization approach in foundation models operating in a zero-shot setting. 
We compare our motif-based tokenization with sample-based tokens in Chronos models \citep{chronos}. 
Additionally, we implement a patch-based version of Chronos, where we alter only the tokenization method and replace the cross-entropy loss function with MSE, which is generally used for continuous patches. We select non-overlapping patches of length \num{4} with similar compression as our high-compression tokenizer and length \num{8}, as recommended in recent literature \citep{goswami2024moment}.
Following these baseline models, we use the T5 architecture \citep{Raffel2020t5} as backbone for our motif-based tokenization and train all models with the same number of tokens, gradient steps, and training settings.
This way, we compare our motif-based tokenization to sample-based tokenization and patches in an isolated setting, ensuring that the tokenization method is the only difference between architectures.
Following Chronos models, we propose models with our tokenizer in \num{5} sizes ranging from tiny (\num{8}\,M parameters) to large (\num{710}\,M parameters).
We evaluate on forecasting \num{64} time series samples, following \citet{chronos}. As context, we utilize \num{128} tokens 
for our and Chronos models and an equivalent input length of \num{384} time series samples for patch-based models.
As literature references, we use patch-based MOMENT \citep{goswami2024moment}, Moirai \citep{woo2024unifiedtraining}, Time-MoE \citep{shi2025timemoe}, and LightGTS \citep{wang2025lightgts} foundation models. 
We restrict evaluation to models with available data and code for reproducibility.

\vspace{0.1\baselineskip}
\section{Results}
\vspace{0.05\baselineskip}
We first demonstrate improvements in forecasting performance and efficiency of our motif-based tokenization over existing methods. Next, we explore the adaptiveness of our tokenizer to diverse temporal patterns of different lengths and complexities and its generalization to unseen data. Finally, we link distinct time series properties, including statistical moments, to our token space. In \cref{appendix:motifs_visualized}, we visualize the learned motifs.

\begin{table*}[!b]
       \small
        \vspace{-0.3\baselineskip}
      \caption{Motif-based tokenization with conditional decoding (cd) and without improves forecasting quality and accelerates models during zero-shot forecasting. We aim for two extremes: best MSE and fastest acceleration. Among  Chronos models, we choose the best as reference. 
      As our tokenization improves MSE while speeding up the model, we are able to choose small models while surpassing the forecasting quality of larger ones. \textbf{Best} in bold.}
      \vspace{-0.5\baselineskip}
      \label{tab:tok_chronos}
      \centering
        \resizebox{0.67\textwidth}{!}{
        \begin{tabular}{lccccc}
            \toprule
            \multirow{2}{*}{Dataset} &  \multicolumn{1}{c}{Chronos} & \multicolumn{2}{c}{Ours} & \multicolumn{2}{c}{Ours$^{\mathrm{cd}}$} \\
            \cmidrule{2-2}\cmidrule{3-4}\cmidrule{5-6}
            & MSE & MSE$^{\mathrm{best}}$ & Accel.$^{\mathrm{fastest}}$ & MSE$^{\mathrm{best}}$ & Accel.$^{\mathrm{fastest}}$\\
            \midrule 
            ETTh1 & \num{0.717} & \num{0.517} & \num{24.88}$\times$ & \textbf{0.459} & \bm{$55.74\times$}\\
            ETTm1 & \num{1.004} & \num{0.637} & \bm{$6.49\times$} & \textbf{0.449} & \bm{$6.49\times$}\\
            Weather & \num{0.265} & \num{0.251} & \num{0.26}$\times$ & \textbf{0.236} & \bm{$3.58\times$}\\
            Electricity & \num{0.222} & \num{0.150} & \bm{$11.20\times$} & \textbf{0.144} & \bm{$11.20\times$}\\
            Traffic & \num{2.717} & \num{0.591} & \bm{$56.66\times$} & \textbf{0.574} & \bm{$56.66\times$}\\
            Solar & \num{1.270} & \num{0.439} & \bm{$4.69\times$} & \textbf{0.371} & \bm{$4.69\times$}\\
            Fev-bench & \num{1.489} & \num{1.006} & \bm{$57.80\times$} & \textbf{0.756} & \bm{$57.80\times$}\\
            \bottomrule 
        \end{tabular}
        }
        \vspace{-8pt}
    \end{table*}

\begin{figure}[t]
    \centering
    \vspace{0.1cm}
    \includegraphics[width=5.5cm,trim={0in 0in 0in 0in},clip]{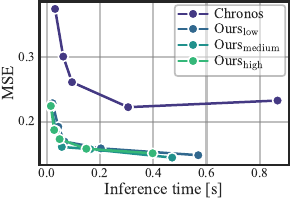}
    \captionsetup{skip=4pt}
    \caption{Zero-shot comparison between our motif-based and sample-wise tokenization (Chronos) on Electricity.}
    \label{fig:tok_electricity}
    \vspace{-1.1\baselineskip}
\end{figure}

\subsection{Efficiency improvements of adaptive tokenization}
\label{sec:tok_vs_singlesample}

Chronos foundation models tokenize every sample of a time series, resulting in many tokens with little temporal information. Especially for large models, this induces substantial computational requirements.  We compare our motif-based tokenization with Chronos models in \num{5} sizes from tiny to large using \num{3} tokenizers (see experimental settings in \cref{sec:experiments}). Chronos and our models are based on the same architecture, training strategy, and dataset. They only \mbox{differ in tokenization.} \\
In our zero-shot evaluation, our motif-based tokenization finds Pareto optimal points on all \num{7} datasets. We show in \cref{fig:tok_electricity} that our tokenizer outperforms Chronos models with sample-based tokenization in forecasting quality and efficiency at the same time. We report our results in \cref{tab:tok_chronos}, choosing the best Chronos model as reference. Among our \num{3} tokenizers and \num{5} model sizes, we illustrate two cases: 1) Selecting the best MSE, 2) Selecting the fastest model that is still better than the Chronos reference. Motif-based tokenization without conditional decoding improves MSE by \SI{39.8}{\percent} and accelerates models \SI{23.14}{\times} on average. With conditional decoding, the improvements are even more substantial, with forecasting quality increasing by \SI{48.0}{\percent} and model acceleration reaching \SI{28.22}{\times}.
On the Traffic dataset, Chronos models diverge during zero-shot testing, while our tokenizer still performs well, highlighting the generalization capability of motifs. 
We show full results in \cref{appendix:tok_single_sample} and further compare our zero-shot motif-based models with state-of-the-art models that are directly trained on the respective datasets. Remarkably, our approach generates the best forecasts in \num{19} out of \num{25} cases without fine-tuning.

\subsection{Comparison with patch-based methods}
\label{sec:tok_vs_patches}
\textbf{Patch-based Chronos}\hspace{1.8mm} Patching, which involves extracting fixed-length subsequences as tokens, compresses the time series and captures local temporal information \citep{nie2023patchtst}. However, patches are rigid and non-adaptive to diverse time series patterns. Here, we compare our adaptive motif-based tokenization with our patch-based Chronos baseline in an isolated setting, where tokenization is the only difference between models. 
Except for ETTh1 and Fev-bench, our tokenization method outperforms all patch-based Chronos models in our isolated comparison in \cref{tab:tok_patches}.
Motif-based tokenization with conditional decoding increases forecasting quality by \SI{12.7}{\percent} on average across all datasets. These results highlight the potential of byte pair encoding for time series.

\textbf{Beyond Chronos}\hspace{1.8mm} We further compare with patch-based literature foundation models MOMENT, Moirai, Time-MoE, and LightGTS in a zero-shot setting. Our tokenizer consistently outperforms all MOMENT and Time-MoE models, and generates better forecasts than LightGTS and Moirai on \num{5} out of \num{7} and \num{4} out of \num{7} datasets, respectively. Given that these models utilize different backbones and datasets, their performance lead in the remaining cases may stem from these systemic advantages rather than the tokenization scheme itself. Nevertheless, these results demonstrate that our tokenizer enables performance competitive with state-of-the-art foundation models.
In \cref{sec:adaptive_compression}, we explore the compression of our motif-based tokenization.

\begin{table*}
      \caption{Benchmarking our motif-based tokenization with conditional decoding (cd) and without against our patch-based Chronos baseline, MOMENT, Moirai, Time-MoE, and LightGTS models, based on zero-shot forecasting quality (MSE). In line with \cref{tab:tok_chronos} we report the best among our tokenizers. We highlight values that are \colorbox{gray!30}{worse} than our method.} 
      \vspace{-0.5\baselineskip}
      \label{tab:tok_patches}
      \centering
        \resizebox{1\textwidth}{!}{
        \begin{tabular}{lrrrrrrrrrrrrrrrrrrrrr}
            \toprule
            \multirow{2}{*}{Dataset} & \multirow{2}{*}{Ours} & \multirow{2}{*}{Ours$^{\mathrm{cd}}$} &\multicolumn{5}{c}{Chronos$^{\mathrm{len=4}}_{\mathrm{patch}}$}&\multicolumn{5}{c}{Chronos$^{\mathrm{len=8}}_{\mathrm{patch}}$} & \multicolumn{3}{c}{MOMENT} & \multicolumn{3}{c}{Moirai}& \multicolumn{2}{c}{Time-MoE} & \multirow{2}{*}{LightGTS} \\\cmidrule{4-8}\cmidrule{9-13}\cmidrule{14-16}\cmidrule{17-19}\cmidrule{20-21} &&&tiny&mini&small&base&large&tiny&mini&small&base&large&small&base&large&small&base&large&base&large&\\\midrule
            ETTh1 & \num{0.517} & \num{0.459}&\cellcolor{gray!30}\num{0.525}&\cellcolor{gray!30}\num{0.474}&\num{0.453}&\cellcolor{gray!30}\num{0.470}&\num{0.384}&\num{0.426}&\num{0.420}&\num{0.446}&\num{0.400}&\num{0.379} & \cellcolor{gray!30}\num{0.765} & \cellcolor{gray!30} \num{0.732} & \cellcolor{gray!30} \num{0.693} & \cellcolor{gray!30} \num{0.465} & \num{0.396} & \num{0.397} &\cellcolor{gray!30}\num{0.767}&\cellcolor{gray!30}\num{0.789}&\num{0.391}\\
            ETTm1 & \num{0.637} & \num{0.449}&\cellcolor{gray!30}\num{0.879}&\cellcolor{gray!30}\num{0.912}&\cellcolor{gray!30}\num{0.916}&\cellcolor{gray!30}\num{1.099}&\cellcolor{gray!30}\num{0.666}&\cellcolor{gray!30}\num{0.800}&\cellcolor{gray!30}\num{0.647}&\cellcolor{gray!30}\num{0.906}&\cellcolor{gray!30}\num{0.704}&\cellcolor{gray!30}\num{0.608} & \cellcolor{gray!30} \num{0.700} & \cellcolor{gray!30} \num{0.710} & \cellcolor{gray!30} \num{0.665} & \cellcolor{gray!30} \num{0.710} &\cellcolor{gray!30} \num{0.600} & \cellcolor{gray!30} \num{0.548} &\cellcolor{gray!30}\num{0.714}&\cellcolor{gray!30}\num{0.733}&\cellcolor{gray!30}\num{0.813}\\
            Weather & \num{0.251} & \num{0.236}&\cellcolor{gray!30}\num{0.425}&\cellcolor{gray!30}\num{0.319}&\cellcolor{gray!30}\num{0.356}&\cellcolor{gray!30}\num{0.601}&\cellcolor{gray!30}\num{0.374}&\cellcolor{gray!30}\num{0.273}&\cellcolor{gray!30}\num{0.305}&\cellcolor{gray!30}\num{0.264}&\cellcolor{gray!30}\num{0.278}&\cellcolor{gray!30}\num{0.284} & \cellcolor{gray!30} \num{0.275} & \cellcolor{gray!30} \num{0.249} & \cellcolor{gray!30} \num{0.240} & \num{0.193} &\num{0.161} & \cellcolor{gray!30} \num{0.245} &\cellcolor{gray!30}\num{0.286}&\cellcolor{gray!30}\num{0.298}&\num{0.157}\\
            Electricity & \num{0.150} & \num{0.144}&\cellcolor{gray!30}\num{0.249}&\cellcolor{gray!30}\num{0.250}&\cellcolor{gray!30}\num{0.227}&\cellcolor{gray!30}\num{0.214}&\cellcolor{gray!30}\num{0.162}&\cellcolor{gray!30}\num{0.220}&\cellcolor{gray!30}\num{0.170}&\cellcolor{gray!30}\num{0.203}&\cellcolor{gray!30}\num{0.169}&\cellcolor{gray!30}\num{0.146} & \cellcolor{gray!30} \num{0.887} & \cellcolor{gray!30} \num{0.888} & \cellcolor{gray!30} \num{0.852} & \cellcolor{gray!30} \num{0.212} & \cellcolor{gray!30} \num{0.163} & \cellcolor{gray!30} \num{0.146} &\cellcolor{gray!30}\num{0.942}&\cellcolor{gray!30}\num{0.944}&\cellcolor{gray!30}\num{0.230}\\
            Traffic & \num{0.591} & \num{0.574}&\cellcolor{gray!30}\num{0.766}&\cellcolor{gray!30}\num{0.808}&\cellcolor{gray!30}\num{0.762}&\cellcolor{gray!30}\num{0.756}&\cellcolor{gray!30}\num{0.624}&\cellcolor{gray!30}\num{0.731}&\cellcolor{gray!30}\num{0.645}&\cellcolor{gray!30}\num{0.685}&\cellcolor{gray!30}\num{0.680}&\cellcolor{gray!30}\num{0.625} & \cellcolor{gray!30} \num{1.458} & \cellcolor{gray!30} \num{1.534} & \cellcolor{gray!30} \num{1.386} & \cellcolor{gray!30} \num{0.645} &\num{0.406} & \num{0.427} &\cellcolor{gray!30}\num{1.571}&\cellcolor{gray!30}\num{1.556}&\cellcolor{gray!30}\num{0.618}\\
            Solar&\num{0.439}&\num{0.371}&\cellcolor{gray!30}\num{1.353}&\cellcolor{gray!30}\num{1.047}&\cellcolor{gray!30}\num{1.004}&\cellcolor{gray!30}\num{1.258}&\cellcolor{gray!30}\num{0.682}&\cellcolor{gray!30}\num{0.856}&\cellcolor{gray!30}\num{0.870}&\cellcolor{gray!30}\num{0.926}&\cellcolor{gray!30}\num{0.841}&\cellcolor{gray!30}\num{0.556}&\cellcolor{gray!30}\num{0.837}&\cellcolor{gray!30}\num{0.832}&\cellcolor{gray!30}\num{0.820}&\cellcolor{gray!30}\num{1.019}&\cellcolor{gray!30}\num{0.947}&\cellcolor{gray!30}\num{1.108}&\cellcolor{gray!30}\num{1.063}&\cellcolor{gray!30}\num{0.959}&\cellcolor{gray!30}\num{0.611}\\
            Fev-bench&\num{1.006}&\num{0.756}&\cellcolor{gray!30}\cellcolor{gray!30}\num{0.869}&\cellcolor{gray!30}\num{0.924}&\cellcolor{gray!30}\num{0.893}&\cellcolor{gray!30}\num{0.952}&\cellcolor{gray!30}\num{0.788}&\cellcolor{gray!30}\num{0.784}&\num{0.720}&\cellcolor{gray!30}\num{0.817}&\cellcolor{gray!30}\num{0.757}&\num{0.680}&\cellcolor{gray!30}\num{1.282}&\cellcolor{gray!30}\num{1.268}&\cellcolor{gray!30}\num{1.240}&\cellcolor{gray!30}\num{0.886}&\cellcolor{gray!30}\num{0.831}&\cellcolor{gray!30}\num{0.785}&\cellcolor{gray!30}\num{1.344}&\cellcolor{gray!30}\num{1.341}&\cellcolor{gray!30}\num{0.918}\\
            \bottomrule 
        \end{tabular}
        }
        \vspace{-1\baselineskip}
    \end{table*}

\subsection{Conditional decoding}
\label{sec:cond_decoding}
\begin{figure}[htbp]
    \centering
    \hfill
    \begin{subfigure}[b]{1.5in}
        \centering
        \includegraphics[width=1.5in]{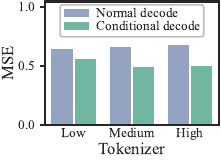}
        \captionsetup{skip=3pt}
        \caption{ETTh1}
    \end{subfigure}
    \hspace{0.3cm}
    \begin{subfigure}[b]{1.5in}
        \centering
        \includegraphics[width=1.5in]{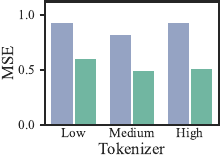}
        \captionsetup{skip=3pt}
        \caption{ETTm1}
    \end{subfigure}
    \hfill
    \vspace{-3pt}
    \caption{Conditional decoding improves forecasting quality for \num{3} tokenizers in small models on \num{2} datasets.}
    \label{fig:cond_decoding}
\end{figure}

Recently emerging foundation models show impressive performance but are expensive to train \citep{chronos}. We propose conditional decoding as a lightweight yet powerful post-hoc optimization method to enhance a model's forecasting quality.  Conditional decoding adds no computational overhead during inference and does not require gradient computation for training. Instead, we analytically compute the global optimum for its few parameters according to \cref{eq:cond_decode_solution}. For our experiments, we utilize \num{3} tokenizers with different compression (see \cref{tab:tokenizers}), \num{7} datasets, and models in size small. In the following, we train conditional decoding to dequantize the models' forecasts on the respective train set and evaluate on the test set. Here, data- and model-dependent conditional decoding may act as an effective domain adaptation method. Later in \cref{appendix:conditional_decoding}, we evaluate data- and model-independent conditional decoding in pure zero-shot settings. \\
Conditional decoding consistently improves forecasting quality in \cref{fig:cond_decoding} in all of our experiments. On the ETTm1 dataset and our tokenizer with high compression, conditional decoding reduces MSE by \SI{44.3}{\percent} with only \num{484} trainable parameters. In \cref{appendix:conditional_decoding}, we provide additional results and further investigate conditional decoding in a data- and model-independent setting. There, conditional decoding mitigates on average \SI{31.9}{\percent} and up to \SI{96.9}{\percent} of our tokenizer's quantization error, enabling us to build tokenizers with even higher compression.

\vspace{-0.1\baselineskip}

\vspace{-0.3\baselineskip}
\subsection{Adaptive compression of diverse time series}
\label{sec:adaptive_compression}

\setlength{\columnsep}{10pt}
   \begin{wraptable}{r}{3.1cm}
      \vspace{-\baselineskip}
      \caption{Average compression of our medium tokenizer on \num{5} datasets.}
      \label{tab:dataset_adaptive_compresison}
      \vspace{-0.5\baselineskip}
      \centering
        \resizebox{3.1cm}{!}{
        \begin{tabular}{lr}
            \toprule 
            Dataset & \multicolumn{1}{c}{$\bar{c}$}\\ 
            \midrule 
            ETTh1 & \num{3.48}\\
            ETTm1 & \num{4.59}\\
            Weather & \num{23.15}\\
            Electricity & \num{3.95}\\
            Traffic & \num{3.30}\\
            \bottomrule 
        \end{tabular}
        }
        \vspace{-1\baselineskip}
    \end{wraptable}

\begin{figure}[htbp]
    \centering
    \hfill
    \begin{subfigure}[b]{1.5in}
        \centering
        \includegraphics[width=1.5in]{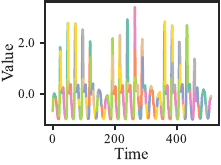}
        \captionsetup{skip=1pt}
        \caption{$\bar{c}=3.56$}
        \label{fig:tokenized_timeseries_a}
    \end{subfigure}
    \hspace{0.3cm}
    \begin{subfigure}[b]{1.5in}
        \centering
        \includegraphics[width=1.5in]{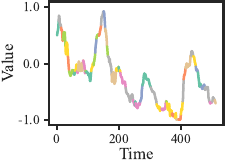}
        \captionsetup{skip=1pt}
        \caption{$\bar{c}=8.13$ }
        \label{fig:tokenized_timeseries_b}
    \end{subfigure}
    \hfill
    \vspace{0.15cm}
    \\
    \hfill
    \begin{subfigure}[b]{1.5in}
        \centering
        \includegraphics[width=1.5in]{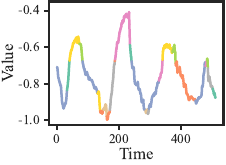}
        \captionsetup{skip=1pt}
        \caption{$\bar{c}=18.95$ }
        \label{fig:tokenized_timeseries_c}
    \end{subfigure}
    \hspace{0.3cm}
    \begin{subfigure}[b]{1.5in}
        \centering
        \includegraphics[width=1.5in]{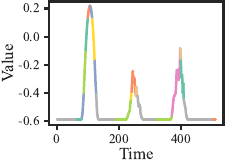}
        \captionsetup{skip=1pt}
        \caption{$\bar{c}=22.26$ }
        \label{fig:tokenized_timeseries_d}
    \end{subfigure}
    \hfill
    \vspace{-3pt}
    \caption{Our adaptive tokenizer \textbf{(a)} exploits periodically recurring motifs on the Traffic dataset and \textbf{(b-d)} compresses time series adaptively depending on pattern complexity on the Weather dataset.}
\end{figure}

Here, we analyze the efficiency benefits of adaptive tokenization in detail. Temporal patterns differ in length and complexity among datasets and within time series \citep{ekambaram2024tinytimemixers, woo2024unifiedtraining}. 
While rigid patches are generally unable to capture these inter- and intra-series variations by employing fixed compression rates, our motif-based tokenization natively exploits these diverse patterns, compressing them adaptively. Here, we analyze our medium compression tokenizer (see \cref{tab:tokenizers}). \\
The Weather dataset contains patterns of various complexities, which we illustrate in \cref{fig:tokenized_timeseries_b,fig:tokenized_timeseries_c,fig:tokenized_timeseries_d}. Here, our tokenizer compresses motifs of different lengths into single tokens, achieving compressions from \num{8.13} up to \num{22.26}. Less complex patterns result in higher compression, while more complex patterns are tokenized more fine-grained. In \cref{fig_appendix:hist_compression}, we demonstrate this adaptive intra-series compression on \num{4} other datasets.
Among datasets, our tokenizer reaches average compressions of \num{3.30} on Traffic and \num{23.15} on Weather in \cref{tab:dataset_adaptive_compresison}. Further, ETTh1 and ETTm1 are sampled with different frequencies but from the same process. The higher compression on ETTm1 indicates that our tokenizer is agnostic to the sampling frequency.
All these results highlight the flexibility of our motif-based tokenization. Compared to MOMENT with patch length \num{8} and \num{7} other patch-based models in \cref{tab:patch_literature_compression_rates}, our tokenizer can yield substantially greater compression by adapting to time series structure.
In \cref{appendix:adaptive_compression}, we further investigate relations between compression of input data and generated tokens and find linear dependencies. We also showcase even higher compressions up to \num{128}. Note that we report efficiency gains in inference time for our main experiments. Here, we investigate adaptive compression of motif-based tokenization at time series level, which directly translates to real-world speed up by requiring fewer autoregressive generation steps. Tokenization overhead is negligible with $<\SI{0.5}{\percent}$ in runtime of our fastest models.

\vspace{0.1\baselineskip}

\vspace{-0.1\baselineskip}
\subsection{Vocabulary complexity and generalization}
\label{sec:vocab_complexity}
Longer motifs benefit the compression and efficiency of our tokenizer. Here, we systematically explore factors influencing the vocabulary complexity and generalization ability. We show that longer motifs are more expressive and enhance forecasting quality. We list further insights in \cref{appendix:pattern_complexity}. In \cref{appendix:noise_robustness}, we investigate robustness to noise, extreme values, and generalization to non-stationary time series.

\textbf{Quantization granularity}\hspace{1.8mm} A lower number of quantization bins $M$ reduces the complexity of the time series, resulting in longer motifs and a smaller vocabulary (see \cref{tab:tokenizers}). However, fewer quantization bins also increase the quantization error, potentially failing to capture important nuances and compromising forecasting quality. In \cref{tab_appendix:main_results,fig_appendix:main_results}, we utilize \num{3} tokenizers with different quantization granularities without conditional decoding, across \num{5} model sizes, and \num{7} datasets to analyze this tradeoff. \\
In \num{25} out of \num{35} settings, our tokenizer with high compression and the largest quantization error leads to the best MSE. This experiment indicates that longer, more expressive motifs benefit forecasting, despite higher quantization error. Moreover, as shown in \cref{sec:cond_decoding}, the quantization error can be largely removed with conditional decoding.

   \begin{wraptable}{r}{3.7cm}
   \vspace{-1\baselineskip}
      \caption{Tokenizers on the Chronos dataset with different token occurrence, vocabulary size, and compression.}
      \vspace{-0.5\baselineskip}
      \label{tab:pmin_tokenizers}
      \centering
        \begin{tabular}{lrr}
            \toprule 
            $p_{\mathrm{min}}$ & \multicolumn{1}{c}{$|\mathcal{V}|$} & \multicolumn{1}{c}{$\bar{c}$}\\ 
            \midrule 
            \num{1000} & \num{1675} & \num{3.18}\\
            \num{8000} & \num{373} & \num{2.50}\\
            \num{32000} & \num{158} & \num{2.08}\\
            \num{128000} & \num{78} & \num{1.66}\\
            \bottomrule 
        \end{tabular}
        \vspace{-\baselineskip}
    \end{wraptable}

\begin{figure}[h]
    \centering
    \begin{subfigure}[b]{1.5in}
        \centering
        \includegraphics[width=1.5in]{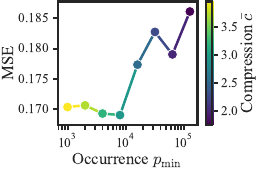}
        \captionsetup{skip=3pt}
        \caption{Electricity}
    \end{subfigure}
    \hspace{0.3cm}
    \begin{subfigure}[b]{1.5in}
        \centering
        \includegraphics[width=1.5in]{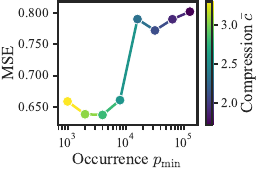}
        \captionsetup{skip=3pt}
        \caption{Traffic}
    \end{subfigure}
    \vspace{-3pt}
    \caption{Varying token occurrence $p_{\mathrm{min}}$ influences forecasting quality for small models. More complex motifs improve MSE.}
    \label{fig:pmin_ablation}
\end{figure}

\textbf{Token occurrence}\hspace{1.8mm}  
There is an inherent tradeoff in tokenization: longer, more complex motifs (created by a high number of recursive merges) naturally occur less frequently in the training data. In the limit, the whole dataset can be represented by a single motif. While setting a lower minimum occurrence threshold $p_{\mathrm{min}}$ allows the vocabulary to capture more complex patterns, these rarer motifs may provide insufficient learning examples for the model to reliably recognize them.
Here, we vary $p_{\mathrm{min}}$ from \num{1000} to \num{128000} training \num{8} different tokenizers. These tokenizers feature different vocabulary complexity and compression, as in \cref{tab:pmin_tokenizers,tab_appendix:pmin_tokenizers}, but have the same quantization error. We base our variations on our medium tokenizer and utilize small models. 
Our results on Electricity and Traffic in \cref{fig:pmin_ablation} indicate an optimal tradeoff. A minimum motif occurrence of $p_{\mathrm{min}}=4000$ times among \num{100}\,M time series samples represents a good balance. Generally, more complex motifs with higher compression result in the best MSE. 

\vspace{-0.2\baselineskip}
\vspace{-0.3\baselineskip}

\textbf{Token level analysis}\hspace{1.8mm} Here, we demonstrate on the token level that complex motifs are a better representation
\vspace*{0.30\baselineskip}
\begin{wraptable}{r}{3.1cm}
        \centering
      \captionof{table}{Correlation $\rho$ of token-level compression and MSE.}
      \vspace{-0.5\baselineskip}
      \label{tab:correlation_compression_mse}
      \centering
        \resizebox{3.25cm}{!}{
        \begin{tabular}{lr}
            \toprule 
            Dataset & \multicolumn{1}{c}{$\rho$}\\ 
            \midrule 
             ETTh1 & \num{-0.26}\\
             ETTm1 & \num{-0.24}\\
             Weather & \num{-0.26}\\
             Electricity & \num{-0.07}\\
             Traffic & \num{-0.27}\\
            \bottomrule 
        \end{tabular}
        }
        \vspace{-\baselineskip}
\end{wraptable}
 for time series generation than their simpler counterparts. To this end, we correlate motif length with token-wise MSE of time series forecasts. We utilize our medium compression tokenizer in a small model. 
On all \num{5} datasets, we observe negative correlation coefficients $\rho$ in \cref{tab:correlation_compression_mse}. Therefore, the generation of longer, more expressive motifs enhances forecasting quality. These results align with our previous findings.

\subsection{Training dataset size}
\label{sec:dataset_size}
   \begin{wraptable}{r}{4.0cm}
      \vspace{-1.0\baselineskip}
      \caption{Influence of training dataset size $N$ on tokenizer vocabulary size, compression, and forecasting quality.}
      \vspace{-0.5\baselineskip}
      \label{tab:training_samples}
      \centering
        \resizebox{3.9cm}{!}{
        \begin{tabular}{lrrr}
            \toprule 
            $N$ & \multicolumn{1}{c}{$|\mathcal{V}|$} & \multicolumn{1}{c}{$\bar{c}$} & MSE\\
            \midrule 
             1\,k & \num{2127} & \num{3.16} & \num{0.569}\\
             10\,k & \num{1853} & \num{3.10} & \num{0.560}\\
             100\,k & \num{1831} & \num{3.11} & \num{0.555}\\
             1\,M & \num{1827} & \num{3.11} & \num{0.533}\\
            \bottomrule 
        \end{tabular}
        }
        \vspace{-1\baselineskip}
    \end{wraptable}
Here, we explore how much data is required to train an efficient motif-based tokenizer. In general, larger datasets better approximate the true distribution of patterns, resulting in more complete vocabularies of motifs $\Psi$. To this end, we train our \num{3} tokenizers on Chronos dataset subsets ranging from \num{1000} to \num{1}\,M time series and scale $p_{\mathrm{min}}$ accordingly. 
Increasing the dataset size improves forecasting quality, as our results in \cref{tab:training_samples} show (averaged across \num{3} tokenizers on \num{5} evaluation datasets). As expected, motifs extracted from a larger sample 
size are less noisy and generalize better. This is also evident in the decreasing 
vocabulary size and compression, indicating a smaller, more universal set of motifs. With \num{1}\,M time series, our tokenizer is highly sample-efficient, requiring less than \SI{10}{\percent} of Chronos data for vocabulary generation. 
We show full results and similar findings for conditional decoding in \cref{appendix:training_set_size}.

\begin{figure*}[!b]
    \centering
    \begin{subfigure}[b]{5.1cm}
        \centering
        \includegraphics[width=5.1cm]{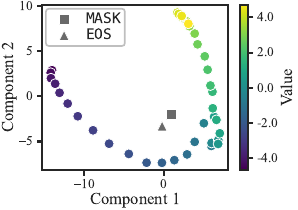}
        \captionsetup{skip=4pt}
        \caption{Symbol values}
    \end{subfigure}
    \hspace{0.3cm}
    \begin{subfigure}[b]{5.1cm}
        \centering
        \includegraphics[width=5.1cm]{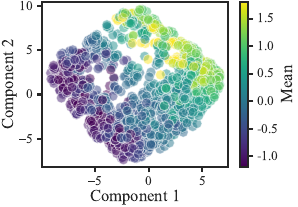}
        \captionsetup{skip=4pt}
        \caption{Motif means}
    \end{subfigure}
    \hspace{0.3cm}
    \begin{subfigure}[b]{5.1cm}
        \centering
        \includegraphics[width=5.1cm]{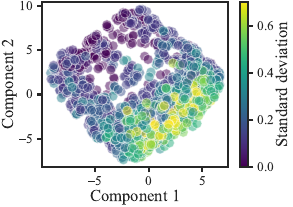}
        \captionsetup{skip=4pt}
        \caption{Motif standard deviations}
    \end{subfigure}
    \vspace{-3pt}
    \caption{Principal component analysis of token embeddings of our medium tokenizer in a small model.}
    \label{fig:pca_e}
\end{figure*}

\subsection{Learned token representations}
\label{sec:explainability_patterns}
Time series have distinct properties such as periodicity, offsets, and trends. A meaningful token representation should model these characteristics. 
Our motif-based tokenization captures periodicity by design, mapping similar patterns at different positions in a time series to the same token. This is qualitatively shown in \cref{fig:tokenized_sequence_zoomed,fig:tokenized_timeseries_a}.
Moreover, we analyze the token embedding space $E$ by doing a principal component analysis in \cref{fig:pca_e,fig_appendix:pca_e}. The learned embeddings successfully capture the values of quantized symbols in $\Omega$ (a), which are separated from \texttt{MASK} and \texttt{EOS} tokens. The embedding space further models the mean (b) and standard deviation (c) of motifs in $\Psi$ in orthogonal dimensions, indicating a good separation of these properties. For motifs with high standard deviation, the model distinguishes between linear and quadratic trends. Finally, motif length is implicitly learned and modeled in the same dimension as the standard deviation, as constant patterns with low standard deviation are likely longer. \\
Our method builds time series motifs hierarchically, where each child token is formed from two parents. Intuitively, a child should be close to its first parent in the embedding space, since the model can predict either the child directly or its parents as a sequence. The average cosine similarity across all tokens is \num{0.072}, while parent–child pairs show a much higher value of \num{0.475}. We illustrate these relations in \cref{fig_appendix:pca_parent_child}, where parents and children are shifted \mbox{along the motif length axis.} \\
In summary, our results confirm that our motif vocabulary yields meaningful time series representations.

\vspace{0.1\baselineskip}
\section{Conclusion}
\label{sec:conclusion}
\vspace{0.05\baselineskip}
In this work, we propose the first pattern-centric tokenization for the time series domain. Our method leverages recurring discrete motifs as tokens and improves forecasting quality and efficiency over existing methods. We further introduce conditional decoding as a lightweight, domain-specific post-hoc optimization method and show its performance gains in large foundation models. We demonstrate our tokenizer's adaptability to patterns of different complexities and show that the learned token embeddings capture meaningful representations of time series properties, including statistical moments and trends.
Finally, our thorough investigation reveals key tradeoffs balancing tokenizer complexity and generalization: discretization granularity presents a dual effect on compression - fewer bins increase discretization error but also make 
patterns more frequent, potentially improving both learnability and compression; 
training data size influences how well the discovered motifs generalize, with smaller datasets being insufficient to learn robust representations of rare motifs. However, with sufficient data, longer and more complex motifs can significantly reduce prediction error, ultimately enhancing compression efficiency. We hope our motif-based tokenization will have a positive effect on reducing the resource consumption and environmental impact of time series models.

\textbf{Limitations}\hspace{1.8mm} 
In our work, we do not conduct hyperparameter search for T5 models due to the high computational cost of training large foundation models. We expect even better results with optimized settings. Moreover, future work can utilize more recent transformer architectures.

\section*{Impact statement}
Motif-based tokenization and conditional decoding demonstrate large accelerations and considerable quality gains throughout a broad range of experiments. We hope that our novel tokenization scheme will contribute to better and more sustainable time series models.

\section*{Disclaimer}
The results, opinions, and conclusions expressed in this publication are not necessarily those of Volkswagen Aktiengesellschaft.

\bibliography{references}

@inproceedings{transformer,
 author = {Vaswani, Ashish and Shazeer, Noam and Parmar, Niki and Uszkoreit, Jakob and Jones, Llion and Gomez, Aidan N and Kaiser, {\L}ukasz and Polosukhin, Illia},
 booktitle = {Advances in Neural Information Processing Systems},
 title = {Attention is All you Need},
 year = {2017}
}

@inproceedings{informer,
  title={Informer: Beyond efficient transformer for long sequence time-series forecasting},
  author={Zhou, Haoyi and Zhang, Shanghang and Peng, Jieqi and Zhang, Shuai and Li, Jianxin and Xiong, Hui and Zhang, Wancai},
  booktitle={AAAI Conference on Artificial Intelligence},
  year={2021}
}

@inproceedings{wu2021autoformer,
 author = {Wu, Haixu and Xu, Jiehui and Wang, Jianmin and Long, Mingsheng},
 booktitle = {Advances in Neural Information Processing Systems},
 title = {Autoformer: Decomposition Transformers with Auto-Correlation for Long-Term Series Forecasting},
 year = {2021}
}

@inproceedings{zhou2022fedformer,
  title={Fedformer: Frequency enhanced decomposed transformer for long-term series forecasting},
  author={Zhou, Tian and Ma, Ziqing and Wen, Qingsong and Wang, Xue and Sun, Liang and Jin, Rong},
  booktitle={International Conference on Machine Learning},
  year={2022},
}

@inproceedings{liu2022non,
 author = {Liu, Yong and Wu, Haixu and Wang, Jianmin and Long, Mingsheng},
 booktitle = {Advances in Neural Information Processing Systems},
 title = {Non-stationary Transformers: Exploring the Stationarity in Time Series Forecasting},
 year = {2022}
}

@inproceedings{
nie2023patchtst,
title={A Time Series is Worth 64 Words: Long-term Forecasting with Transformers},
author={Yuqi Nie and Nam H Nguyen and Phanwadee Sinthong and Jayant Kalagnanam},
booktitle={International Conference on Learning Representations},
year={2023},
}

@inproceedings{
liu2022pyraformer,
title={Pyraformer: Low-Complexity Pyramidal Attention for Long-Range Time Series Modeling and Forecasting},
author={Shizhan Liu and Hang Yu and Cong Liao and Jianguo Li and Weiyao Lin and Alex X. Liu and Schahram Dustdar},
booktitle={International Conference on Learning Representations},
year={2022},
}

@article{cirstea2022triformer,
      title={Triformer: Triangular, Variable-Specific Attentions for Long Sequence Multivariate Time Series Forecasting--Full Version}, 
      author={Razvan-Gabriel Cirstea and Chenjuan Guo and Bin Yang and Tung Kieu and Xuanyi Dong and Shirui Pan},
      year={2022},
      eprint={2204.13767},
      journal={arXiv:2204.13767},
      archivePrefix={arXiv},
      primaryClass={cs.LG}
}

@inproceedings{
zhang2023crossformer,
title={Crossformer: Transformer Utilizing Cross-Dimension Dependency for Multivariate Time Series Forecasting},
author={Yunhao Zhang and Junchi Yan},
booktitle={International Conference on Learning Representations},
year={2023},
}

@article{garza2023timegpt1,
      title={TimeGPT-1}, 
      author={Azul Garza and Max Mergenthaler-Canseco},
      year={2023},
      eprint={2310.03589},
     journal={arXiv:2310.03589}  ,
      archivePrefix={arXiv},
      primaryClass={cs.LG}
}

@article{das2024timesfm,
      title={A decoder-only foundation model for time-series forecasting}, 
      author={Abhimanyu Das and Weihao Kong and Rajat Sen and Yichen Zhou},
      year={2023},
      eprint={2310.10688},
     journal={arXiv:2310.10688}   ,
      archivePrefix={arXiv},
      primaryClass={cs.CL}
}

@article{rasul2024lagllama,
      title={Lag-Llama: Towards Foundation Models for Probabilistic Time Series Forecasting}, 
      author={Kashif Rasul and Arjun Ashok and Andrew Robert Williams and Hena Ghonia and Rishika Bhagwatkar and Arian Khorasani and Mohammad Javad Darvishi Bayazi and George Adamopoulos and Roland Riachi and Nadhir Hassen and Marin Biloš and Sahil Garg and Anderson Schneider and Nicolas Chapados and Alexandre Drouin and Valentina Zantedeschi and Yuriy Nevmyvaka and Irina Rish},
      year={2023},
     journal={arXiv:2310.08278} ,
      eprint={2310.08278},
      archivePrefix={arXiv},
      primaryClass={cs.LG}
}

@article{woo2024unifiedtraining,
      title={Unified Training of Universal Time Series Forecasting Transformers}, 
      author={Gerald Woo and Chenghao Liu and Akshat Kumar and Caiming Xiong and Silvio Savarese and Doyen Sahoo},
      year={2024},
      eprint={2402.02592},
     journal={arXiv:2402.02592}   ,
      archivePrefix={arXiv},
      primaryClass={cs.LG}
}

@inproceedings{
dosovitskiy2021vit,
title={An Image is Worth 16x16 Words: Transformers for Image Recognition at Scale},
author={Alexey Dosovitskiy and Lucas Beyer and Alexander Kolesnikov and Dirk Weissenborn and Xiaohua Zhai and Thomas Unterthiner and Mostafa Dehghani and Matthias Minderer and Georg Heigold and Sylvain Gelly and Jakob Uszkoreit and Neil Houlsby},
booktitle={International Conference on Learning Representations},
year={2021},
}

@inproceedings{chronos,
      title={Chronos: Learning the Language of Time Series}, 
      author={Abdul Fatir Ansari and Lorenzo Stella and Caner Turkmen and Xiyuan Zhang and Pedro Mercado and Huibin Shen and Oleksandr Shchur and Syama Sundar Rangapuram and Sebastian Pineda Arango and Shubham Kapoor and Jasper Zschiegner and Danielle C. Maddix and Hao Wang and Michael W. Mahoney and Kari Torkkola and Andrew Gordon Wilson and Michael Bohlke-Schneider and Yuyang Wang},
      year={2024},
      booktitle = {Transactions on Machine Learning Research}
}

@inproceedings{
godahewa2021monash,
title={Monash Time Series Forecasting Archive},
author={Rakshitha Wathsadini Godahewa and Christoph Bergmeir and Geoffrey I. Webb and Rob Hyndman and Pablo Montero-Manso},
booktitle={Neural Information Processing Systems Datasets and Benchmarks Track},
year={2021},
}

@InProceedings{kingma2015adam,
  author={Kingma, Diederik P. and Ba, Jimmy L.},
  booktitle = {International Conference on Learning Representations},
  title     = {Adam: A Method for Stochastic Optimization},
  year      = {2015},
}

@inproceedings{wang2024timexer,
 author = {Yuxuan Wang and Haixu Wu and Jiaxiang Dong and Guo Qin and Haoran Zhang and Yong Liu and Yunzhong Qiu and Jianmin Wang and Mingsheng Long},
 booktitle = {Advances in Neural Information Processing Systems},
 title = {TimeXer: Empowering Transformers for Time Series Forecasting with Exogenous Variables},
 year = {2024}
}

@inproceedings{wu2024perimidformer,
title={Peri-midFormer: Periodic Pyramid Transformer for Time Series Analysis},
author={Qiang Wu and Gechang Yao and Zhixi Feng and Shuyuan Yang},
booktitle={Advances in Neural Information Processing Systems},
year={2024},
}

@inproceedings{
masserano2024wavelettok,
title={Enhancing Foundation Models for Time Series Forecasting via Wavelet-based Tokenization},
author={Luca Masserano and Abdul Fatir Ansari and Boran Han and Xiyuan Zhang and Christos Faloutsos and Michael W. Mahoney and Andrew Gordon Wilson and Youngsuk Park and Syama Sundar Rangapuram and Danielle C. Maddix and Bernie Wang},
booktitle={International Conference on Machine Learning},
year={2025},
}

@inproceedings{
goswami2024moment,
title={{MOMENT}: A Family of Open Time-series Foundation Models},
author={Mononito Goswami and Konrad Szafer and Arjun Choudhry and Yifu Cai and Shuo Li and Artur Dubrawski},
booktitle={International Conference on Machine Learning},
year={2024},
}

@inproceedings{ekambaram2024tinytimemixers,
 author = {Ekambaram, Vijay and Jati, Arindam and Dayama, Pankaj and Mukherjee, Sumanta and Nguyen, Nam H. and Gifford, Wesley M. and Reddy, Chandra and Kalagnanam, Jayant},
 booktitle = {Advances in Neural Information Processing Systems},
 title = {Tiny Time Mixers (TTMs): Fast Pre-trained Models for Enhanced Zero/Few-Shot Forecasting of Multivariate Time Series},
 year = {2024}
}

@inproceedings{
liu2024timer,
title={Timer: Generative Pre-trained Transformers Are Large Time Series Models},
author={Yong Liu and Haoran Zhang and Chenyu Li and Xiangdong Huang and Jianmin Wang and Mingsheng Long},
booktitle={International Conference on Machine Learning},
year={2024},
}

@inproceedings{gao2024units,
 author = {Gao, Shanghua and Koker, Teddy and Queen, Owen and Hartvigsen, Thomas and Tsiligkaridis, Theodoros and Zitnik, Marinka},
 booktitle = {Advances in Neural Information Processing Systems},
 title = {UniTS: A Unified Multi-Task Time Series Model},
 year = {2024}
}

@inproceedings{torgo1997regressiontoclassification,
title = {Regression using classification algorithms},
booktitle = {Intelligent Data Analysis},
year = {1997},
author = {Luís Torgo and João Gama},
}

@inproceedings{oord2017vqvae,
 author = {van den Oord, Aaron and Vinyals, Oriol and kavukcuoglu, koray},
 booktitle = {Advances in Neural Information Processing Systems},
 title = {Neural Discrete Representation Learning},
 year = {2017}
}

@inproceedings{
talukder2024totem,
title={{TOTEM}: {TO}kenized Time Series {EM}beddings for General Time Series Analysis},
author={Sabera J Talukder and Yisong Yue and Georgia Gkioxari},
booktitle={Transactions on Machine Learning Research},
year={2024},
}

@inproceedings{chen2024sdformer,
 author = {Chen, Zhicheng and Feng, Shibo and Zhang, Zhong and Xiao, Xi and Gao, Xingyu and Zhao, Peilin},
 booktitle = {Advances in Neural Information Processing Systems},
 title = {SDformer: Similarity-driven Discrete Transformer For Time Series Generation},
 year = {2024}
}

@article{elsner2024multidimensionalbytepairencoding,
      title={Multidimensional Byte Pair Encoding: Shortened Sequences for Improved Visual Data Generation}, 
      author={Tim Elsner and Paula Usinger and Julius Nehring-Wirxel and Gregor Kobsik and Victor Czech and Yanjiang He and Isaak Lim and Leif Kobbelt},
      year={2024},
      eprint={2411.10281},
      archivePrefix={arXiv},
      primaryClass={cs.CV},
      journal={arXiv:2411.10281},
}

@inproceedings{gage1994bpeoriginal,
author = {Gage, Philip},
title = {A new algorithm for data compression},
year = {1994},
booktitle = {The C Users Journal},
}

@techreport{shibata1999bpetextcompression,
author={Shibata, Yusuxke and Kida, Takuya and Fukamachi, Shuichi and Takeda, Masayuki and Shinohara, Ayumi and Shinohara, Takeshi and Arikawa, Setsuo},
year = {1999},
title = {Byte Pair Encoding: A Text Compression Scheme That Accelerates Pattern Matching},
institution = {Department of Informatics, Kyushu University, Japan},
}

@inproceedings{sennrich2016bpetokenization,
title = {Neural Machine Translation of Rare Words with Subword Units},
author = {Rico Sennrich and Barry Haddow and Alexandra Birch},
year = {2016},
booktitle = {Annual Meeting of the Association for Computational Linguistics},
}

@inproceedings{
sommer2023powerofmotifs,
title={The Power of Motifs as Inductive Bias for Learning Molecular Distributions},
author={Johanna Sommer and Leon Hetzel and David L{\"u}dke and Fabian J Theis and Stephan G{\"u}nnemann},
booktitle={ICLR 2023 - Machine Learning for Drug Discovery workshop},
year={2023},
}

@inproceedings{Raffel2020t5,
author = {Raffel, Colin and Shazeer, Noam and Roberts, Adam and Lee, Katherine and Narang, Sharan and Matena, Michael and Zhou, Yanqi and Li, Wei and Liu, Peter J.},
title = {Exploring the limits of transfer learning with a unified text-to-text transformer},
year = {2020},
booktitle = {Journal of Machine Learning Research},
}

@inproceedings{
goetz2025localmerging,
title={Efficient Time Series Processing for Transformers and State-Space Models through Token Merging},
author={Leon Götz and Marcel Kollovieh and Stephan Günnemann and Leo Schwinn},
booktitle={International Conference on Machine Learning},
year={2025},
}

@inproceedings{
liu2025sundial,
title={Sundial: A Family of Highly Capable Time Series Foundation Models},
author={Yong Liu and Guo Qin and Zhiyuan Shi and Zhi Chen and Caiyin Yang and Xiangdong Huang and Jianmin Wang and Mingsheng Long},
booktitle={International Conference on Machine Learning},
year={2025},
}

@inproceedings{
chen2025acloserlook,
title={A Closer Look at Transformers for Time Series Forecasting: Understanding Why They Work and Where They Struggle},
author={Yu Chen and Nathalia C{\'e}spedes and Payam Barnaghi},
booktitle={International Conference on Machine Learning},
year={2025},
}

@inproceedings{lin2003sax,
  title={A symbolic representation of time series, with implications for streaming algorithms},
  author={Lin, Jessica and Keogh, Eamonn and Lonardi, Stefano and Chiu, Bill},
  booktitle={8th ACM SIGMOD workshop on Research issues in data mining and knowledge discovery},
  year={2003}
}

@inproceedings{berndt1994timeseries_dtw,
  title={Using dynamic time warping to find patterns in time series},
  author={Berndt, Donald J and Clifford, James},
  booktitle={International conference on knowledge discovery and data mining},
  year={1994}
}

@article{cohen2024toto,
      title={Toto: Time Series Optimized Transformer for Observability}, 
      author={Ben Cohen and Emaad Khwaja and Kan Wang and Charles Masson and Elise Ramé and Youssef Doubli and Othmane Abou-Amal},
      year={2024},
      eprint={2407.07874},
      archivePrefix={arXiv},
      primaryClass={cs.LG},
      journal={arXiv:2407.07874},
}

@article{auer2025tirex,
      title={TiRex: Zero-Shot Forecasting Across Long and Short Horizons with Enhanced In-Context Learning}, 
      author={Andreas Auer and Patrick Podest and Daniel Klotz and Sebastian Böck and Günter Klambauer and Sepp Hochreiter},
      year={2025},
      eprint={2505.23719},
      archivePrefix={arXiv},
      primaryClass={cs.LG},
      journal={arXiv:2505.23719},
}

@article{liu2024moiraimoe,
      title={Moirai-MoE: Empowering Time Series Foundation Models with Sparse Mixture of Experts}, 
      author={Xu Liu and Juncheng Liu and Gerald Woo and Taha Aksu and Yuxuan Liang and Roger Zimmermann and Chenghao Liu and Silvio Savarese and Caiming Xiong and Doyen Sahoo},
      year={2024},
      eprint={2410.10469},
      archivePrefix={arXiv},
      primaryClass={cs.LG},
      journal={arXiv:2410.10469},
}

@inproceedings{shi2025timemoe,
  title={Time-moe: Billion-scale time series foundation models with mixture of experts},
  author={Shi, Xiaoming and Wang, Shiyu and Nie, Yuqi and Li, Dianqi and Ye, Zhou and Wen, Qingsong and Jin, Ming},
  booktitle={International conference on learning representations},
  year={2025}
}

@inproceedings{wang2025lightgts,
title={Light{GTS}: A Lightweight General Time Series Forecasting Model},
author={Yihang Wang and Yuying Qiu and Peng Chen and Yang Shu and Zhongwen Rao and Lujia Pan and Bin Yang and Chenjuan Guo},
booktitle={International Conference on Machine Learning},
year={2025},
}

@article{shchur2026fevbench,
      title={fev-bench: A Realistic Benchmark for Time Series Forecasting}, 
      author={Oleksandr Shchur and Abdul Fatir Ansari and Caner Turkmen and Lorenzo Stella and Nick Erickson and Pablo Guerron and Michael Bohlke-Schneider and Yuyang Wang},
      year={2025},
      eprint={2509.26468},
      archivePrefix={arXiv},
      primaryClass={cs.LG},
      journal={2509.26468},
}
\bibliographystyle{icml2026}

\newpage
\appendix
\onecolumn
\crefalias{section}{appendix}
\crefalias{subsection}{appendix}

\section{An adaptive tokenization approach for time series}
\label{appendix:algorithms}
We provide pseudocode for generating a vocabulary of motifs and utilizing the motifs to tokenize a time series. 

\begin{algorithm}
\caption{Motif vocabulary generation according to \cref{eq:motif_vocabulary}.}
\begin{algorithmic}
\Require Dataset of discretized time series $\mathcal{D^\prime}$, minimum motif occurrence $p_{\mathrm{min}}$
\Ensure Motif vocabulary $\Psi$
\State $\Psi \gets \{\}$ \Comment{Initialize empty vocabulary}
\State $z^\prime_{\mathrm{new}} \gets M+2$ \Comment{Account for quantized symbols and $\{\texttt{MASK}, \texttt{EOS}\}$}
\While{true} \Comment{Iteratively find motifs}
\State $\mathrm{pair,cnt} \gets $ count $(z^\prime_i, z^\prime_{i+1})$ in $\mathcal{D^\prime}$ \Comment{Most frequent adjacent token pair and its count}
\If{$\mathrm{cnt} \geq p_{\mathrm{min}}$}
\State $z^\prime_{\mathrm{new}} \gets z^\prime_{\mathrm{new}}+1$ \Comment{Allocate new token ID}
\State $\Psi[\mathrm{pair}] \gets z^\prime_{\mathrm{new}}$ \Comment{Add new token to vocabulary $(z^\prime_i, z^\prime_{i+1}) \rightarrow z^\prime_{\mathrm{new}}$}
\State $\mathcal{D^\prime} \gets \mathcal{D^\prime}\setminus\{\mathrm{pair}\} \cup \{z^\prime_{\mathrm{new}}\}$ \Comment{Replace new token in dataset $\mathcal{D^\prime}$}
\Else
\State \Return $\Psi$ \Comment{Token occurs to infrequent}
\EndIf
\EndWhile

\end{algorithmic}
\end{algorithm}

\begin{algorithm}
\caption{Tokenization of a discretized time series according to \cref{eq:tokenization_m}.}
\begin{algorithmic}
\Require Discretized time series $z^\prime$, motif vocabulary $\Psi$
\Ensure Tokenized time series $z^{\prime\prime}$
\For{$\psi$ \textbf{in} $\Psi$}\Comment{Iterate over motifs in order of vocabulary creation}
    \State $\psi_{\mathrm{key}}, \psi_{\mathrm{value}} \gets \psi$ \Comment{$\psi$ made of key value mappings $(z^\prime_i, z^\prime_{i+1})\rightarrow z^\prime_{\mathrm{new}}$} 
    \For{$(z^\prime_i, z^\prime_{i+1})$ \textbf{in} $z^\prime$} \Comment{Iterate over adjacent tokens in $z^\prime$}
        \If{$(z^\prime_i, z^\prime_{i+1})$ matches $\psi_{\mathrm{key}}$} \Comment{Adjacent tokens match motif}
        \State replace $(z^\prime_i, z^\prime_{i+1})$ with $\psi_{\mathrm{value}}$ in $z^\prime$ \Comment{Replace tokens with motif: shortens $z^\prime$ by $1$}
        \EndIf
    \EndFor
\EndFor
\State $z^{\prime\prime} \gets z^\prime$
\State \Return $z^{\prime\prime}$
\end{algorithmic}
\end{algorithm}

\newpage

\section{Experiments}
\label{appendix:experiments}
In this section, we list additional information about our experimental settings and resources.

\textbf{Datasets}\hspace{1.8mm} We train our models and tokenizers on the recently proposed Chronos dataset \citep{chronos}. It contains \num{11}\,M time series with over \num{11}\,B samples. Time series are curated from \num{28} real-world datasets or are generated synthetically. Due to its diverse nature and size, this dataset is well-suited for training foundation models. \\
We base our zero-shot evaluation on \num{6} commonly used time series datasets and the Fev-bench benchmark covering different forecasting applications: 
\textit{ETTh1} and \textit{ETTm1} measure the power load and temperature of electric transformers in hourly and quarter-hourly granularity~\citep{informer}.
\textit{Weather} consists of meteorological quantities such as air temperature and is recorded every $10$ minutes in $2020$.\footnote{\url{https://www.bgc-jena.mpg.de/wetter/}}
\textit{Electricity} measures the energy demand of households in hourly granularity~\citep{godahewa2021monash}.
\textit{Traffic} consists of hourly road occupancies in the San Francisco Bay Area~\citep{godahewa2021monash}.
\textit{Solar} measures the power production of photovoltaic plants in \num{10} minute intervals~\citep{godahewa2021monash}. \textit{Fev-bench} is a broad time series forecasting benchmark composed of datasets from diverse domains \citep{shchur2026fevbench}. We exclude sequences with missing values and those shorter than our input and prediction horizons. In total, we evaluate on \num{37} datasets from Fev-bench spanning \num{5} domains, as in \cref{tab:fev_bench_datasets}, and report averaged results.

\vspace{-0.2cm}
\begin{table}[H]
  \caption{Datasets from the Fev-bench benchmark used in our experiments.}
  \vspace{-0.5\baselineskip}
  \label{tab:fev_bench_datasets}
  \centering
  \resizebox{0.75\columnwidth}{!}{
  \begin{tabular}{lc}
    \toprule
    Domain & Dataset\\
    \midrule 
    \multirow{4}{*}{Cloud} & BizITObs L2C 5T, BizITObs L2C 1H, BOOMLET 619, BOOMLET 772, \\
    & BOOMLET 963, BOOMLET 1209, BOOMLET 1225, BOOMLET 1230, \\
    & BOOMLET 1282, BOOMLET 1487, BOOMLET 1631, BOOMLET 1676, \\
    & BOOMLET 1855, BOOMLET 2187, Redset 5T, Redset 1H \\
    \midrule 
    Economy & US Consumption 1M \\
    \midrule 
    \multirow{3}{*}{Energy} &  ENTSO-e Load 15T, ENTSO-e Load 30T, ENTSO-e Load 1H, EPF-BE, \\
    & EPF-DE, EPF-FR, EPF-NP, EPF-PJM, ERCOT 1D, ERCOT 1W,\\
    & ETT 1H, GFC12, GFC14, GFC17, Solar with Weather 1H \\
    \midrule 
    Mobility & Loop Seattle 5t, Loop Seattle 1H, M-DENSE 1H, SZ Taxi 15T \\
    \midrule 
    Retail & M5 1D \\
    \bottomrule
  \end{tabular}
  }
\end{table}

\textbf{Hyperparameters}\hspace{1.8mm} For training our T5 models, we utilize the hyperparameters of Chronos \citep{chronos}, which we list in \cref{tab_appendix:hyperparameters}. We expect even better results of our tokenizer when performing hyperparameter tuning. However, this is very expensive for large foundation models.

  \vspace{-0.2cm}
   \begin{table}[H]
      \caption{Hyperparameters of our T5 backbone models in \num{5} sizes from tiny to large.}
      \vspace{-0.5\baselineskip}
      \label{tab_appendix:hyperparameters}
      \centering
        \begin{tabular}{lccccc}
            \toprule 
            Hyperparameter & tiny & mini & small & base & large\\ 
            \midrule 
            \textbf{T5 models} & & & & & \\
            Token dimension $d$ & \num{256}& \num{384}& \num{512}& \num{768}& \num{1024}\\
            Encoder layers & \num{4}& \num{4}& \num{6}& \num{12}& \num{24}\\
            Decoder layers & \num{4}& \num{4}& \num{6}& \num{12}& \num{24}\\
            Heads & \num{4}& \num{8}& \num{8}& \num{12}& \num{16}\\
            \midrule 
            \textbf{Training} & & & & & \\
            Seed & \multicolumn{5}{c}{\num{2024}} \\
            Activation & \multicolumn{5}{c}{ReLU} \\
            Dropout rate & \multicolumn{5}{c}{\num{0.1}} \\
            Learning rate  & \multicolumn{5}{c}{\num{0.001}} \\
            Learning rate decay  & \multicolumn{5}{c}{linear} \\
            Gradient steps  & \multicolumn{5}{c}{\num{200000}} \\
            Batch size  & \multicolumn{5}{c}{\num{256}} \\
            Optimizer  & \multicolumn{5}{c}{Adam~\citep{kingma2015adam}} \\
            
            \bottomrule 
        \end{tabular}
    \end{table}

\newpage

\textbf{Reproducibility of measurements}\hspace{1.8mm} In our zero-shot evaluations, we use the same data splits as \citet{wu2021autoformer}. We evaluate once and report results on the test set. \\
Regarding \textbf{predictive quality}, we report MSE standard deviations for our most common experimental settings here, as repeating all experiments multiple times is computationally infeasible for large foundation models. To this end, we train small models with \num{6} randomized seeds with our medium compression tokenizer and Chronos baselines with sample-based tokenization and patches of length \num{8}. Our zero-shot evaluation on \num{5} datasets in \cref{appendix_tab:mse_std} demonstrates low average MSE standard deviations of \SI{6.7}{\percent} for our motif-based tokenization, \SI{4.2}{\percent} for Chronos baselines with sample-based tokenization, and \SI{10.1}{\percent} for patch-based Chronos models, supporting the significance of our results. \\
For our main experiments, where we compare models of different sizes, we report the end-to-end \textbf{inference time} as it is of high practical interest. This also includes tokenization and detokenization overhead, which is negligible in practice with $<\SI{0.5}{\percent}$ in runtime of our fastest models. We use the same Nvidia A6000 GPU for profiling with \num{2} warm-up and \num{2} measurement runs per batch to achieve inference time standard deviations $< \SI{2}{\percent}$.\\
Regarding efficiency measures, we additionally report the \textbf{compression} at time series level of our tokenizer. This is a hardware- and model-independent measure and the metric most related to our work. Needing to process fewer tokens or requiring fewer autoregressions directly translates to improvements in inference time of models, which, however, is a hardware-dependent measure.\\
Finally, we suggest executing tokenization and detokenization as pipelined pre- and postprocessing operations on the CPU. This way, the minimal tokenization overhead does not affect throughput at all as model execution on the \mbox{GPU is the limiting factor.}

\begin{table}[H]
      \caption{MSE standard deviations for small models with our medium compression tokenizer and Chronos baselines on \num{5} datasets computed from \num{6} random seeds.} 
      \vspace{-0.5\baselineskip}
      \label{appendix_tab:mse_std}
      \centering
        \begin{tabular}{lccc}
            \toprule
            \multirow{2}{*}{Dataset} &  \multicolumn{2}{c}{Chronos} & \multirow{2}{*}{Ours} \\
            \cmidrule{2-3}
            & sample-based & patch & \\
            \midrule 
            ETTh1 & \num{0.009} & \num{0.022} & \num{0.130}  \\
            ETTm1 & \num{0.036} & \num{0.134} & \num{0.051}  \\
            Weather & \num{0.017} & \num{0.026} & \num{0.011}  \\
            Electricity & \num{0.007} & \num{0.026} & \num{0.003}  \\
            Traffic & \num{0.228} & \num{0.027} & \num{0.017}  \\
            \bottomrule 
        \end{tabular}
    \end{table}

\textbf{Computational effort}\hspace{1.8mm} Building the vocabulary of our tokenizer is an iterative process. Computationally, this is rather cheap and we execute it on a single core of an Intel Xeon w5-3435X CPU. For our medium compression tokenizer and \num{100000} time series with a total of \num{100}\,M samples, vocabulary generation only takes \num{3.8} hours utilizing \SI{1.2}{\giga\byte} of CPU memory. Analytically computing the conditional decoding distributions is even faster and generally takes under \num{10} seconds.\\
For training the T5 models, we utilize Nvidia H100 GPUs. In total, we train \num{36} foundation models of different sizes and with different motif-based tokenizers. 
We estimate the computational effort to reproduce our experiments in \cref{tab:computehours}. Please note that we reuse previously trained tokenizers and models in most of our experiments.

\begin{table}[H]
  \caption{Computational effort to reproduce our experiments.}
  \vspace{-0.5\baselineskip}
  \label{tab:computehours}
  \centering
  \begin{tabular}{lrr}
    \toprule
    Experiment & Device & Hours  \\
    \midrule
    \textbf{Tokenizer} & &\\
    low & CPU & \num{9.1}\\
    medium & CPU & \num{3.8} \\
    high & CPU & \num{2.5} \\
    \midrule
    \textbf{T5 models} & &\\
    Chronos baselines & GPU & \num{4350} \\
    Main experiments & GPU & \num{4800} \\
    Vocabulary complexity and generalization & GPU & \num{2240} \\
    Training dataset size & GPU & \num{2880} \\
    MSE standard deviations & GPU & \num{4050} \\
    
    \bottomrule
  \end{tabular}
\end{table}

\newpage

\section{Results}
Here, we show additional experiments and results.

\subsection{Preprocessing strategies}
\label{appendix:preprocessing_strategies}
\begin{figure}[h]
    \centering
    \includegraphics[width=2.8in,trim={0in 0in 0in 0in},clip]{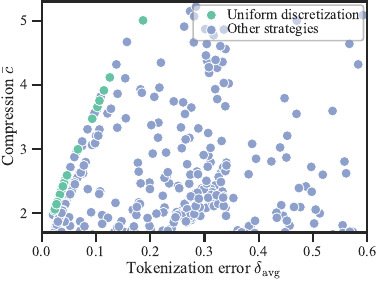}
    \vspace{-3pt}
    \caption{Comparison of uniform discretization with other tokenization preprocessing strategies, including derivatives, Gaussian smoothing, window-based normalization, and Gaussian- and data-distribution-based discretization.}
    \label{fig_appendix:preprocessing_strategies}
\end{figure}

We conduct new experiments exploring different preprocessing strategies before discretizing a time series. Each of these methods features different tradeoffs between signal preservation and noise rejection. The first derivative of a time series $z$ removes its offset, potentially yielding more similar motifs. However, derivatives generally introduce noise. To counter this, we utilize Gaussian kernels to smooth the time series. We further employ window-based normalization. Besides uniform distributions for discretization, we experiment with Gaussian distributions and the precise data distribution $P(\mathcal{D})$. \\
We conduct an extensive search among combinations of preprocessing strategies on \num{500} tokenizers in \cref{fig_appendix:preprocessing_strategies}. Uniform discretization with a different number of bins $M$ is Pareto optimal in balancing the average tokenization error $\delta_{\mathrm{avg}}$ and compression $\bar{c}$. We utilize this method throughout our paper.

\vspace{5pt}

\newpage

\subsection{Efficiency improvements of adaptive tokenization}
\label{appendix:tok_single_sample}
In \cref{tab_appendix:main_results,fig_appendix:main_results}, we report full, non-aggregated results comparing our motif-based tokenization with Chronos foundation models that tokenize every sample. We conduct additional experiments to compare our motif-based tokenization with non-foundation models tokenizing every sample. Finally, we isolate the effects of discretization and temporal motif representation and investigate different input lengths.

\vspace{0.3cm}
\begin{table}[H]
      \caption{Comparison of MSE and inference time of Chronos models and our low, medium, and high compression tokenizers with conditional decoding (cd) and without on \num{7} datasets and \num{5} model sizes. \textbf{Best} MSE in bold.} 
      \vspace{-0.5\baselineskip}
      \label{tab_appendix:main_results}
      \centering
\resizebox{1\textwidth}{!}{
\begin{tabular}{lrrrrrrrrrrrr}\toprule
\multirow{2}{*}{Dataset}&\multirow{2}{*}{Model size}&\multicolumn{2}{c}{Chronos}&\multicolumn{3}{c}{Ours$_{\mathrm{low}}$}&\multicolumn{3}{c}{Ours$_{\mathrm{medium}}$}&\multicolumn{3}{c}{Ours$_{\mathrm{high}}$}\\
\cmidrule{3-4}\cmidrule{5-7}\cmidrule{8-10}\cmidrule{11-13}&&MSE&time&MSE&MSE$^{\mathrm{cd}}$&time&MSE&MSE$^{\mathrm{cd}}$&time&MSE&MSE$^{\mathrm{cd}}$&time\\
\midrule\multirow{5}{*}{ETTh1}&tiny&\num{0.744}&\SI{0.031}{\second}&\num{0.854}&\num{0.617}&\SI{0.022}{\second}&\num{0.720}&\num{0.545}&\SI{0.016}{\second}&\num{0.881}&\textbf{0.540}&\SI{0.016}{\second}\\
&mini&\num{0.736}&\SI{0.061}{\second}&\num{0.803}&\num{0.599}&\SI{0.044}{\second}&\num{0.585}&\textbf{0.507}&\SI{0.035}{\second}&\num{0.758}&\num{0.520}&\SI{0.031}{\second}\\
&small&\num{0.741}&\SI{0.094}{\second}&\num{0.656}&\num{0.565}&\SI{0.067}{\second}&\num{0.669}&\textbf{0.500}&\SI{0.057}{\second}&\num{0.686}&\num{0.512}&\SI{0.051}{\second}\\
&base&\num{0.759}&\SI{0.305}{\second}&\num{0.602}&\num{0.525}&\SI{0.204}{\second}&\num{0.554}&\num{0.465}&\SI{0.175}{\second}&\num{0.528}&\textbf{0.463}&\SI{0.165}{\second}\\
&large&\num{0.717}&\SI{0.867}{\second}&\num{0.530}&\num{0.487}&\SI{0.575}{\second}&\num{0.527}&\num{0.461}&\SI{0.507}{\second}&\num{0.517}&\textbf{0.459}&\SI{0.456}{\second}\\
\midrule\multirow{5}{*}{ETTm1}&tiny&\num{1.138}&\SI{0.031}{\second}&\num{1.044}&\num{0.637}&\SI{0.020}{\second}&\num{1.063}&\num{0.619}&\SI{0.016}{\second}&\num{0.904}&\textbf{0.585}&\SI{0.014}{\second}\\
&mini&\num{1.105}&\SI{0.061}{\second}&\num{1.031}&\num{0.644}&\SI{0.041}{\second}&\num{1.018}&\textbf{0.560}&\SI{0.033}{\second}&\num{1.017}&\num{0.589}&\SI{0.030}{\second}\\
&small&\num{1.004}&\SI{0.094}{\second}&\num{0.934}&\num{0.609}&\SI{0.064}{\second}&\num{0.826}&\textbf{0.495}&\SI{0.054}{\second}&\num{0.933}&\num{0.520}&\SI{0.050}{\second}\\
&base&\num{1.061}&\SI{0.305}{\second}&\num{0.887}&\num{0.590}&\SI{0.184}{\second}&\num{0.759}&\num{0.473}&\SI{0.165}{\second}&\num{0.660}&\textbf{0.460}&\SI{0.152}{\second}\\
&large&\num{1.084}&\SI{0.867}{\second}&\num{0.764}&\num{0.569}&\SI{0.540}{\second}&\num{0.784}&\num{0.487}&\SI{0.488}{\second}&\num{0.637}&\textbf{0.449}&\SI{0.438}{\second}\\
\midrule\multirow{5}{*}{Weather}&tiny&\num{0.313}&\SI{0.031}{\second}&\num{0.525}&\num{0.331}&\SI{0.015}{\second}&\num{0.406}&\num{0.290}&\SI{0.012}{\second}&\num{0.338}&\textbf{0.284}&\SI{0.013}{\second}\\
&mini&\num{0.297}&\SI{0.061}{\second}&\num{0.482}&\num{0.305}&\SI{0.032}{\second}&\num{0.324}&\textbf{0.257}&\SI{0.026}{\second}&\num{0.313}&\num{0.280}&\SI{0.027}{\second}\\
&small&\num{0.265}&\SI{0.094}{\second}&\num{0.463}&\num{0.298}&\SI{0.050}{\second}&\num{0.344}&\num{0.250}&\SI{0.046}{\second}&\num{0.290}&\textbf{0.238}&\SI{0.044}{\second}\\
&base&\num{0.266}&\SI{0.305}{\second}&\num{0.535}&\num{0.316}&\SI{0.138}{\second}&\num{0.307}&\textbf{0.241}&\SI{0.140}{\second}&\num{0.273}&\num{0.258}&\SI{0.132}{\second}\\
&large&\num{0.269}&\SI{0.867}{\second}&\num{0.492}&\num{0.316}&\SI{0.418}{\second}&\num{0.293}&\num{0.242}&\SI{0.405}{\second}&\num{0.251}&\textbf{0.236}&\SI{0.367}{\second}\\
\midrule\multirow{5}{*}{Electricity}&tiny&\num{0.375}&\SI{0.031}{\second}&\num{0.246}&\num{0.228}&\SI{0.021}{\second}&\num{0.241}&\textbf{0.223}&\SI{0.016}{\second}&\num{0.245}&\num{0.224}&\SI{0.015}{\second}\\
&mini&\num{0.301}&\SI{0.061}{\second}&\num{0.200}&\num{0.192}&\SI{0.043}{\second}&\num{0.198}&\textbf{0.186}&\SI{0.033}{\second}&\num{0.199}&\num{0.187}&\SI{0.027}{\second}\\
&small&\num{0.261}&\SI{0.094}{\second}&\num{0.176}&\num{0.169}&\SI{0.066}{\second}&\num{0.170}&\textbf{0.161}&\SI{0.056}{\second}&\num{0.185}&\num{0.173}&\SI{0.048}{\second}\\
&base&\num{0.222}&\SI{0.305}{\second}&\num{0.167}&\num{0.159}&\SI{0.203}{\second}&\num{0.165}&\textbf{0.157}&\SI{0.163}{\second}&\num{0.166}&\num{0.158}&\SI{0.148}{\second}\\
&large&\num{0.233}&\SI{0.867}{\second}&\num{0.154}&\num{0.148}&\SI{0.569}{\second}&\num{0.150}&\textbf{0.144}&\SI{0.471}{\second}&\num{0.158}&\num{0.151}&\SI{0.397}{\second}\\
\midrule\multirow{5}{*}{Traffic}&tiny&\num{4.682}&\SI{0.031}{\second}&\num{0.805}&\num{0.756}&\SI{0.021}{\second}&\num{0.825}&\num{0.762}&\SI{0.017}{\second}&\num{0.755}&\textbf{0.721}&\SI{0.015}{\second}\\
&mini&\num{3.751}&\SI{0.061}{\second}&\num{0.716}&\num{0.684}&\SI{0.042}{\second}&\num{0.682}&\textbf{0.648}&\SI{0.033}{\second}&\num{0.680}&\num{0.650}&\SI{0.027}{\second}\\
&small&\num{2.722}&\SI{0.094}{\second}&\num{0.693}&\num{0.646}&\SI{0.065}{\second}&\num{0.659}&\textbf{0.617}&\SI{0.056}{\second}&\num{0.646}&\num{0.627}&\SI{0.047}{\second}\\
&base&\num{3.413}&\SI{0.305}{\second}&\num{0.686}&\num{0.631}&\SI{0.201}{\second}&\num{0.630}&\textbf{0.585}&\SI{0.163}{\second}&\num{0.631}&\num{0.608}&\SI{0.143}{\second}\\
&large&\num{2.717}&\SI{0.867}{\second}&\num{0.688}&\num{0.628}&\SI{0.576}{\second}&\num{0.613}&\num{0.576}&\SI{0.474}{\second}&\num{0.591}&\textbf{0.574}&\SI{0.386}{\second}\\
\midrule\multirow{5}{*}{Solar}&tiny&\num{1.387}&\SI{0.031}{\second}&\num{1.633}&\num{0.729}&\SI{0.014}{\second}&\num{1.348}&\textbf{0.595}&\SI{0.013}{\second}&\num{1.117}&\num{0.613}&\SI{0.013}{\second}\\
&mini&\num{1.270}&\SI{0.061}{\second}&\num{1.442}&\num{0.715}&\SI{0.029}{\second}&\num{0.963}&\num{0.521}&\SI{0.028}{\second}&\num{0.767}&\textbf{0.491}&\SI{0.027}{\second}\\
&small&\num{1.358}&\SI{0.094}{\second}&\num{1.316}&\num{0.720}&\SI{0.045}{\second}&\num{0.845}&\textbf{0.449}&\SI{0.048}{\second}&\num{0.677}&\num{0.450}&\SI{0.045}{\second}\\
&base&\num{1.311}&\SI{0.305}{\second}&\num{1.265}&\num{0.710}&\SI{0.123}{\second}&\num{0.870}&\textbf{0.458}&\SI{0.131}{\second}&\num{0.688}&\num{0.491}&\SI{0.125}{\second}\\
&large&\num{1.319}&\SI{0.867}{\second}&\num{1.210}&\num{0.715}&\SI{0.388}{\second}&\num{0.751}&\num{0.419}&\SI{0.419}{\second}&\num{0.493}&\textbf{0.371}&\SI{0.369}{\second}\\
\midrule\multirow{5}{*}{Fev-bench}&tiny&\num{1.562}&\SI{0.031}{\second}&\num{1.223}&\num{0.892}&\SI{0.021}{\second}&\num{1.195}&\num{0.862}&\SI{0.017}{\second}&\num{1.093}&\textbf{0.815}&\SI{0.015}{\second}\\
&mini&\num{1.499}&\SI{0.061}{\second}&\num{1.155}&\num{0.864}&\SI{0.043}{\second}&\num{1.137}&\num{0.835}&\SI{0.034}{\second}&\num{1.089}&\textbf{0.799}&\SI{0.029}{\second}\\
&small&\num{1.584}&\SI{0.094}{\second}&\num{1.119}&\num{0.839}&\SI{0.066}{\second}&\num{1.082}&\num{0.807}&\SI{0.094}{\second}&\num{1.069}&\textbf{0.778}&\SI{0.049}{\second}\\
&base&\num{1.577}&\SI{0.305}{\second}&\num{1.095}&\num{0.832}&\SI{0.189}{\second}&\num{1.071}&\num{0.780}&\SI{0.164}{\second}&\num{1.038}&\textbf{0.769}&\SI{0.141}{\second}\\
&large&\num{1.489}&\SI{0.867}{\second}&\num{1.060}&\num{0.802}&\SI{0.556}{\second}&\num{1.033}&\num{0.778}&\SI{0.475}{\second}&\num{1.006}&\textbf{0.756}&\SI{0.391}{\second}\\
\bottomrule\end{tabular}
}
    \end{table}

\newpage

\begin{figure}[H]
    \centering
    \begin{subfigure}[b]{2.2in}
        \centering
        \includegraphics[width=2.2in]{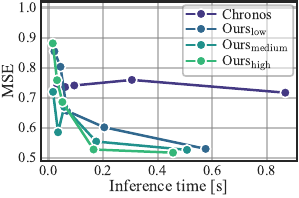}
        \captionsetup{skip=1pt}
        \caption{ETTh1, without conditional decoding}
    \end{subfigure}
    \hspace{1cm}
        \begin{subfigure}[b]{2.2in}
        \centering
        \includegraphics[width=2.2in]{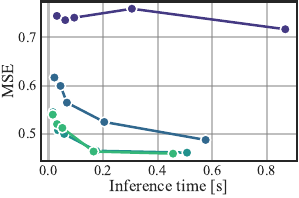}
        \captionsetup{skip=1pt}
        \caption{ETTh1, with conditional decoding}
    \end{subfigure}
    
    \vspace{0.2cm}
    \begin{subfigure}[b]{2.2in}
        \centering
        \includegraphics[width=2.2in]{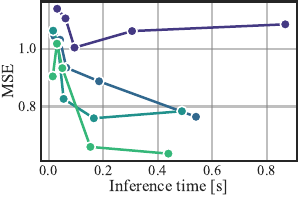}
        \captionsetup{skip=1pt}
        \caption{ETTm1, without conditional decoding}
    \end{subfigure}
    \hspace{1cm}
        \begin{subfigure}[b]{2.2in}
        \centering
        \includegraphics[width=2.2in]{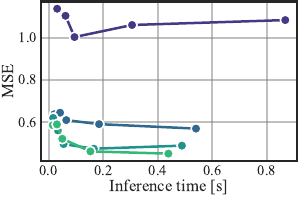}
        \captionsetup{skip=1pt}
        \caption{ETTm1, with conditional decoding}
    \end{subfigure}
    
    \vspace{0.2cm}
    \begin{subfigure}[b]{2.2in}
        \centering
        \includegraphics[width=2.2in]{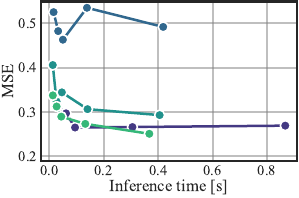}
        \captionsetup{skip=1pt}
        \caption{Weather, without conditional decoding}
    \end{subfigure}
    \hspace{1cm}
        \begin{subfigure}[b]{2.2in}
        \centering
        \includegraphics[width=2.2in]{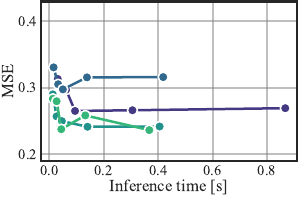}
        \captionsetup{skip=1pt}
        \caption{Weather, with conditional decoding}
    \end{subfigure}
    
    \vspace{0.2cm}
    \begin{subfigure}[b]{2.2in}
        \centering
        \includegraphics[width=2.2in]{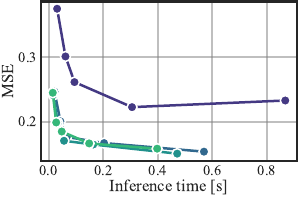}
        \captionsetup{skip=1pt}
        \caption{Electricity, without conditional decoding}
    \end{subfigure}
    \hspace{1cm}
        \begin{subfigure}[b]{2.2in}
        \centering
        \includegraphics[width=2.2in]{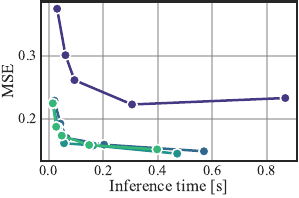}
        \captionsetup{skip=1pt}
        \caption{Electricity, with conditional decoding}
    \end{subfigure}
    
    \vspace{0.2cm}
    \begin{subfigure}[b]{2.2in}
        \centering
        \includegraphics[width=2.2in]{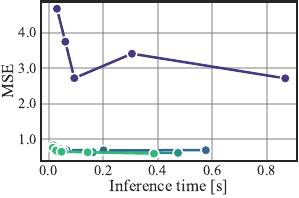}
        \captionsetup{skip=1pt}
        \caption{Traffic, without conditional decoding}
    \end{subfigure}
    \hspace{1cm}
        \begin{subfigure}[b]{2.2in}
        \centering
        \includegraphics[width=2.2in]{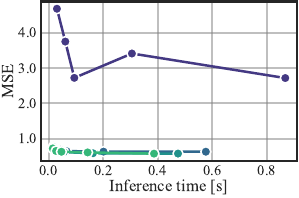}
        \captionsetup{skip=1pt}
        \caption{Traffic, with conditional decoding}
    \end{subfigure}
    \vspace{-3pt}
    \caption{Comparison of our motif-based tokenization with and without conditional decoding with Chronos models tokenizing every sample during zero-shot evaluation on \num{7} datasets and \num{5} model sizes.}
    \label{fig_appendix:main_results}
\end{figure}

\begin{figure}[t]\ContinuedFloat
    \centering
    \begin{subfigure}[b]{2.2in}
        \centering
        \includegraphics[width=2.2in]{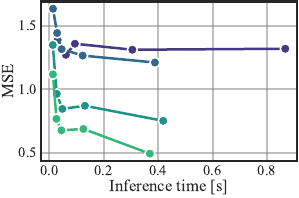}
        \captionsetup{skip=1pt}
        \caption{Solar, without conditional decoding}
    \end{subfigure}
    \hspace{1cm}
    \begin{subfigure}[b]{2.2in}
        \centering
        \includegraphics[width=2.2in]{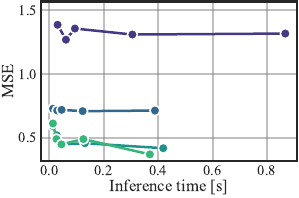}
        \captionsetup{skip=1pt}
        \caption{Solar, with conditional decoding}
    \end{subfigure}

    \vspace{0.2cm}
    \begin{subfigure}[b]{2.2in}
        \centering
        \includegraphics[width=2.2in]{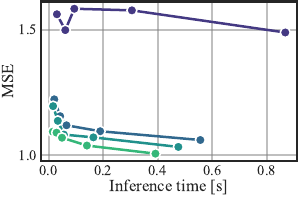}
        \captionsetup{skip=1pt}
        \caption{Fev-bench, without conditional decoding}
    \end{subfigure}
    \hspace{1cm}
    \begin{subfigure}[b]{2.2in}
        \centering
        \includegraphics[width=2.2in]{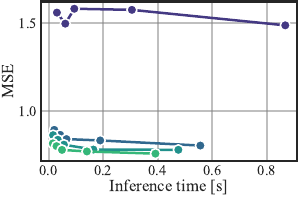}
        \captionsetup{skip=1pt}
        \caption{Fev-bench, with conditional decoding}
    \end{subfigure}

    \captionsetup{list=no}
    \caption{(continued)}
\end{figure}

\textbf{Non-foundation models}\hspace{1.8mm} Here, we compare our motif-based tokenization in T5 foundation models with non-foundation models that are specifically trained on the ETTh1, ETTm1, Weather, Electricity, and Traffic datasets. We utilize common time series architectures including Autoformer \citep{wu2021autoformer}, FEDformer \citep{zhou2022fedformer}, Informer \citep{informer}, Non-stationary \citep{liu2022non}, and vanilla transformers \citep{transformer}. These models extract tokens as a multivariate slice for every time step. For comparison with our motif-based tokenization, we utilize the results of \citet{goetz2025localmerging} for \num{2} layer models in \cref{appendix_tab:non_fm}. The authors forecast \num{96} time series samples from \num{192} context tokens.\\
In \num{19} out of \num{25} cases, our foundation model in a zero-shot setting outperforms the specifically trained models in forecasting quality.

\begin{table}[H]
      \caption{Comparison of our motif-based tokenization with conditional decoding (cd) and without in zero-shot foundation models with non-foundation models that tokenize every sample, based on forecasting quality (MSE). We highlight values that are \colorbox{gray!30}{worse} than our method.} 
      \vspace{-0.5\baselineskip}
      \label{appendix_tab:non_fm}
      \centering
        \begin{tabular}{lccccccc}
            \toprule
            Dataset & Ours & Ours$^{\mathrm{cd}}$ &  Autoformer & FEDformer & Informer & Non-stationary & Transformer \\
            \midrule 
            ETTh1 & \num{0.52} & \num{0.46} & \num{0.42} & \num{0.38} & \cellcolor{gray!30}\num{0.87} & \cellcolor{gray!30}\num{0.55} & \cellcolor{gray!30}\num{0.75} \\
            ETTm1 & \num{0.64} & \num{0.45} & \num{0.44} & \num{0.36} & \cellcolor{gray!30}\num{0.65} & \num{0.42} & \cellcolor{gray!30}\num{0.52} \\
            Weather & \num{0.25} & \num{0.24} & \cellcolor{gray!30}\num{0.28} & \cellcolor{gray!30}\num{0.27} & \cellcolor{gray!30}\num{0.35} & \num{0.19} & \cellcolor{gray!30}\num{0.25} \\
            Electricity & \num{0.15} & \num{0.14} & \cellcolor{gray!30}\num{0.18} & \cellcolor{gray!30}\num{0.20} & \cellcolor{gray!30}\num{0.30} & \cellcolor{gray!30}\num{0.17} & \cellcolor{gray!30}\num{0.25} \\
            Traffic & \num{0.59} & \num{0.57} & \cellcolor{gray!30}\num{0.63} & \cellcolor{gray!30}\num{0.59} & \cellcolor{gray!30}\num{0.68} & \cellcolor{gray!30}\num{0.60} & \cellcolor{gray!30}\num{0.66} \\
            \bottomrule 
        \end{tabular}
    \end{table}

\newpage

\textbf{Effects of discretization and temporal motifs}\hspace{1.8mm} Our motif-based tokenization consists of two steps: Quantizing the time series into a sequence of discrete symbols and compressing this sequence using a vocabulary of temporal motifs. To isolate the contributions of discretization and motif representation, we train models from tiny to large with the same number of quantization bins ($M=37$) as our medium compression tokenizer, but without applying motif discovery.\\
Our results in \cref{tab:discretization_motifs} show that models relying solely on discretization perform worse than those incorporating motif-based representations, except for the Weather dataset. However, they still outperform Chronos baselines. 
The improved MSE of the motif-based approach can be attributed to the fact that longer motifs provide the model with higher-level building blocks, making sequence prediction easier and more accurate (this is supported by our results that longer motifs are associated with smaller MSE, see \cref{sec:vocab_complexity}). Moreover, our motif representation also compresses temporal patterns, resulting in efficiency gains, which is a main motivation for our work. Please note that the Weather dataset features extraordinarily high average compressions of \num{23.15} in \cref{tab:dataset_adaptive_compresison}, which may be the reason for the decreased accuracy. However, here the motif-based model is substantially more efficient.

\begin{table}[H]
       \small
      \caption{Forecasting quality (MSE) on \num{5} evaluation datasets for models from tiny to large, with $M=37$ quantization bins, trained with and without motif representations. \textbf{Best} in bold.}
      \vspace{-0.5\baselineskip}
      \label{tab:discretization_motifs}
      \centering
        \begin{tabular}{lcccccc}
            \toprule
            \multirow{2}{*}{Dataset} &  \multicolumn{5}{c}{Discretization} & \multirow{2}{*}{Motifs} \\
            \cmidrule{2-6}
            & tiny & mini & small & base & large\\
            \midrule 
            ETTh1 & \num{0.611} & \num{0.608} &\num{0.600} &\num{0.625} &\num{0.626} &\textbf{0.526}\\
            ETTm1 & \num{1.035} & \num{0.911} &\num{0.928} &\num{0.887} &\num{0.918} &\textbf{0.759}\\
            Weather & \num{0.231} & \num{0.229} &\num{0.220} &\num{0.222} &\textbf{0.215} &\num{0.293}\\
            Electricity & \num{0.252} & \num{0.223} &\num{0.211} &\num{0.196} &\num{0.189} &\textbf{0.150}\\
            Traffic & \num{1.099} & \num{1.135} &\num{1.195} &\num{1.122} &\num{1.184} &\textbf{0.613}\\
            \bottomrule 
        \end{tabular}
    \end{table}

\textbf{Effects of input length}\hspace{1.8mm} In our main experiments, we compare sample-based tokenization in Chronos models with our motif-based tokenizer. We use \num{128} context tokens in both cases to ensure an identical compute budget for the transformer encoder, such that any efficiency gains arise solely from compressed token generation in the decoder. As a consequence, the two tokenization schemes induce different effective input lengths. To disentangle the effect of input length from token expressivity, we additionally compare sample-based tokenization to our medium compression tokenizer without conditional decoding using the same input length of \num{384} time series samples. Note that all comparisons to patch-based tokenizers also use this context length. \\
\Cref{tab:different_input_length} confirms that the observed improvements in predictive performance result from the increased expressiveness of motif-based tokenization rather than from longer inputs. Due to the larger input length, these Chronos models are on average \num{1.24}$\times$ slower than those reported throughout. As noted earlier, the Weather dataset exhibits exceptionally high average compression ratios of \num{23.15} in \cref{tab:dataset_adaptive_compresison}, which may explain the reduced accuracy. However, the motif-based model remains substantially more efficient in this setting.

\begin{table}[H]
      \caption{Forecasting quality (MSE) on \num{5} evaluation datasets for Chronos models and our motif-based tokenization using the same input length of \num{384} time series samples. \textbf{Best} in bold.}
      \vspace{-0.5\baselineskip}
      \label{tab:different_input_length}
      \centering
        \begin{tabular}{lcccccc}
            \toprule
            \multirow{2}{*}{Dataset} &  \multicolumn{5}{c}{Chronos} & \multirow{2}{*}{Motifs} \\
            \cmidrule{2-6}
            & tiny & mini & small & base & large\\
            \midrule 
            ETTh1 & \num{0.697} & \num{0.713} &\num{0.693} &\num{0.630} &\num{0.636} &\textbf{0.527}\\
            ETTm1 & \num{1.059} & \num{0.999} &\num{0.862} &\num{0.787} &\num{0.764} &\textbf{0.759}\\
            Weather & \num{0.263} & \textbf{0.223} &\num{0.241} &\num{0.233} &\num{0.253} &\num{0.293}\\
            Electricity & \num{0.312} & \num{0.266} &\num{0.208} &\num{0.184} &\num{0.172} &\textbf{0.150}\\
            Traffic & \num{3.296} & \num{2.793} &\num{2.549} &\num{1.934} &\num{1.937} &\textbf{0.613}\\
            \bottomrule 
        \end{tabular}
    \end{table}

\newpage

\subsection{Conditional decoding}
\label{appendix:conditional_decoding}

In this section, we provide full results on data- and model-dependent conditional decoding trained on the models' predictions in \cref{appendix_fig:cond_decode}. We further explore conditional decoding in a data- and model-independent setting. Additionally, we investigate higher-order conditional decoding schemes and demonstrate that conditional decoding can also enhance the predictive performance of patch-based models.
\vspace{0.3cm}

\begin{figure}[H]
    \centering
    \hfill
    \begin{subfigure}[b]{2.2in}
        \centering
        \includegraphics[width=2.2in]{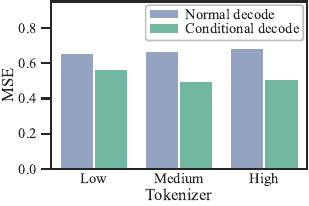}
        \captionsetup{skip=3pt}
        \caption{ETTh1}
    \end{subfigure}
    \hspace{1.5cm}
    \begin{subfigure}[b]{2.2in}
        \centering
        \includegraphics[width=2.2in]{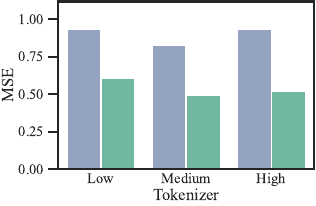}
        \captionsetup{skip=3pt}
        \caption{ETTm1}
    \end{subfigure}
    \hfill
    \vspace{0.4cm}
    \\
    \hfill
    \begin{subfigure}[b]{2.2in}
        \centering
        \includegraphics[width=2.2in]{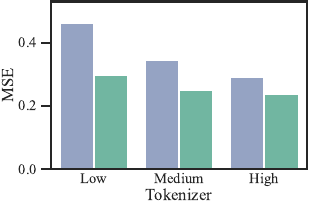}
        \captionsetup{skip=3pt}
        \caption{Weather}
    \end{subfigure}
    \hspace{1.5cm}
    \begin{subfigure}[b]{2.2in}
        \centering
        \includegraphics[width=2.2in]{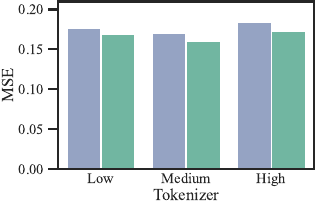}
        \captionsetup{skip=3pt}
        \caption{Electricity}
    \end{subfigure}
    \hfill
    \vspace{0.4cm}
    \\
    \hfill
    \begin{subfigure}[b]{2.2in}
        \centering
        \includegraphics[width=2.2in]{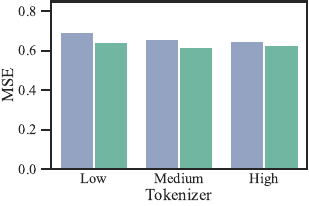}
        \captionsetup{skip=3pt}
        \caption{Traffic}
    \end{subfigure}
    \hspace{1.5cm}
    \begin{subfigure}[b]{2.2in}
        \centering
        \includegraphics[width=2.2in]{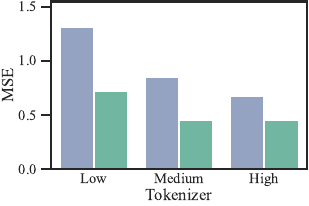}
        \captionsetup{skip=3pt}
        \caption{Solar}
    \end{subfigure}
    \hfill
    \vspace{0.4cm}
    \\
    \begin{subfigure}[b]{2.2in}
        \centering
        \includegraphics[width=2.2in]{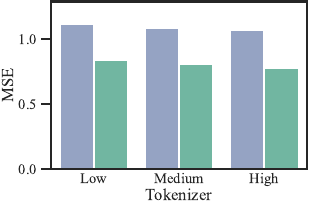}
        \captionsetup{skip=3pt}
        \caption{Fev-bench}
    \end{subfigure}
    \vspace{-3pt}
    \caption{Conditional decoding improves forecasting quality for \num{3} tokenizers in small models on \num{7} datasets.}
    \label{appendix_fig:cond_decode}
\end{figure}

\newpage

\textbf{Data- and model-independent conditional decoding}\hspace{1.8mm} Here, we investigate conditional decoding in a data- and model-independent setting to universally improve the forecasting quality of foundation models. To this end, we train conditional decoding to dequantize quantized time series $z^\prime = \mathbf{q}_\Omega(z)$. Here, conditional decoding is model-independent and can only mitigate the quantization error as this is the only error introduced. This is why we report MSE improvements relative to the average quantization error $\delta_{\mathrm{avg}}$ on the respective evaluation datasets. We utilize the Chronos dataset for training and \num{5} datasets for zero-shot evaluation. We demonstrate conditional decoding on our \num{3} tokenizers and small models. \\
In \num{14} out of \num{15} settings in \cref{tab_appendix:cond_decode_zero_shot}, conditional decoding improves forecasting quality. On ETTm1, it mitigates up to \SI{96.9}{\percent} of the quantization error of our tokenizer with high compression. This enables us to build tokenizers with even higher compression and quantization error, as it can be effectively recovered.

\begin{table}[H]
      \caption{Conditional decoding in data- and model-independent setting recovers the quantization errors of our \num{3} tokenizers by varying degrees on \num{5} datasets.}
      \vspace{-0.5\baselineskip}
      \label{tab_appendix:cond_decode_zero_shot}
      \centering
        \begin{tabular}{lrrr}
            \toprule 
            \multirow{2}{*}{Dataset} &  \multicolumn{3}{c}{Compression}  \\
            \cmidrule{2-4}
             & \multicolumn{1}{c}{low} & \multicolumn{1}{c}{medium} & \multicolumn{1}{c}{high}\\ 
            \midrule 
            ETTh1 & \SI{22.0}{\percent} & \SI{21.3}{\percent} & \SI{49.4}{\percent}\\
            ETTm1 & \SI{61.1}{\percent} & \SI{16.1}{\percent} & \SI{96.9}{\percent}\\
            Weather & \SI{87.6}{\percent} & \SI{38.9}{\percent} & \SI{25.0}{\percent}\\
            Electricity & \SI{13.5}{\percent} & \SI{23.0}{\percent} & \SI{21.4}{\percent}\\
            Traffic & \SI{0.0}{\percent} & \SI{0.7}{\percent} & \SI{1.7}{\percent}\\
            \bottomrule 
        \end{tabular}
    \end{table}

\textbf{Higher-order conditional decoding}\hspace{1.8mm} We propose conditional decoding as a post-hoc optimization method for transforming discrete tokens back into continuous space. Throughout our experiments, we explore very lightweight conditional decoding using the first-order Markov assumption. Here, we investigate longer look-back windows, i.e., conditioning on the \num{2} or \num{3} previous samples. These higher-order models with exponentially more parameters may lead to better performance in special cases but might also hinder generalization or overfit. We utilize small models, our medium compression tokenizer, and data- and model-dependent conditional decoding (see \cref{sec:cond_decoding}) for this experiment. \\
Our results in \cref{appendix_tab:higher_order_cd} demonstrate that higher-order conditional decoding only marginally improves predictive performance. While first-order conditional decoding has $M^2=$ \num{1369} parameters, parameter count is substantially increased for third-order models $M^4=$ \num{1874161}.
This demonstrates the effectiveness of our first-order method.

\begin{table}[H]
      \caption{Comparison of conditional decoding with different Markov orders on \num{7} datasets. \textbf{Best} in bold.
      } 
      \vspace{-0.5\baselineskip}
      \label{appendix_tab:higher_order_cd}
      \centering
        \begin{tabular}{lcccc}
            \toprule
            \multirow{2}{*}{Dataset} & \multirow{2}{*}{MSE} & \multicolumn{3}{c}{MSE$^{\mathrm{cd}}$} \\
            \cmidrule{3-5}
            && first-order & second-order & third-order\\
            \midrule 
            ETTh1 & \num{0.669} & \textbf{0.500} & \num{0.503}& \num{0.516}  \\
            ETTm1 & \num{0.826} & \num{0.495} & \num{0.489}& \textbf{0.486}  \\
            Weather & \num{0.344} & \num{0.250} & \num{0.246}& \textbf{0.242}  \\
            Electricity & \num{0.170} & \textbf{0.161} & \textbf{0.161}& \num{0.163}  \\
            Traffic & \num{0.659} & \textbf{0.617} & \textbf{0.617}& \num{0.623}  \\
            Solar & \num{0.845} & \num{0.449} & \num{0.447}& \textbf{0.445}  \\
            Fev-bench & \num{1.082} & \num{0.807} & \num{0.775}& \textbf{0.738}  \\
            \midrule 
            Average & \num{0.656} & \num{0.468} & \num{0.463}& \num{0.459}  \\
            \bottomrule 
        \end{tabular}
    \end{table}

\newpage

\textbf{Conditional decoding for patch-based models}\hspace{1.8mm} We demonstrate a direct application of conditional decoding to post-hoc refine the predictions of patch‑based models from the literature. To this end, we first quantize the predictions of patch‑based models to reduce noise. In a second step, we transform them back into time series representations using data- and model-dependent conditional decoding (see \cref{sec:cond_decoding}). Here, conditional decoding may simultaneously act as a lightweight domain adaptation, promoting dataset-specific patterns.\\
Our results in \cref{appendix_tab:tok_patches} show that conditional decoding improves the forecasting quality of patch‑based models by \SI{5}{\percent} on average, without modifying or retraining the model itself. This post‑hoc application makes our tokenization directly compatible with current continuous‑embedding architectures in the literature.

\begin{table}[H]
      \caption{Tokenization of predictions of patch-based literature models with subsequent conditional decoding post-hoc improves prediction quality by \SI{5.0}{\percent} on average without modifying or retraining the model itself.} 
      \vspace{-0.5\baselineskip}
      \label{appendix_tab:tok_patches}
      \centering
        \resizebox{0.9\textwidth}{!}{
        \begin{tabular}{lrrrrrrrrr}
            \toprule
            \multirow{2}{*}{Dataset}  & \multicolumn{3}{c}{MOMENT} & \multicolumn{3}{c}{Moirai}& \multicolumn{2}{c}{Time-MoE} & \multirow{2}{*}{LightGTS}\\\cmidrule{2-4}\cmidrule{5-7}\cmidrule{8-9}&small&base&large&small&base&large&base&large\\\midrule
            ETTh1 & \SI{0.0}{\percent} & \SI{0.0}{\percent} & \SI{0.0}{\percent} & \SI{0.0}{\percent} & \SI{0.0}{\percent} & \SI{0.0}{\percent} & \SI{0.2}{\percent} & \SI{0.4}{\percent} & \SI{0.8}{\percent} \\
            ETTm1 & \SI{0.7}{\percent} & \SI{0.0}{\percent} & \SI{0.0}{\percent} & \SI{5.8}{\percent} & \SI{3.2}{\percent} & \SI{1.3}{\percent} & \SI{0.0}{\percent} & \SI{0.8}{\percent} & \SI{10.9}{\percent} \\
            Weather & \SI{0.4}{\percent} & \SI{0.0}{\percent} & \SI{0.0}{\percent} & \SI{1.6}{\percent} & \SI{0.0}{\percent} & \SI{25.7}{\percent} & \SI{1.4}{\percent} & \SI{4.3}{\percent} & \SI{0.0}{\percent} \\
            Electricity & \SI{2.6}{\percent} & \SI{0.0}{\percent} & \SI{0.0}{\percent} & \SI{2.3}{\percent} & \SI{0.0}{\percent} & \SI{0.0}{\percent} & \SI{5.6}{\percent} & \SI{4.9}{\percent} & \SI{6.2}{\percent} \\
            Traffic & \SI{3.5}{\percent} & \SI{2.4}{\percent} & \SI{0.4}{\percent} & \SI{0.0}{\percent} & \SI{0.0}{\percent} & \SI{0.0}{\percent} & \SI{6.0}{\percent} & \SI{4.3}{\percent} & \SI{0.0}{\percent} \\
            Solar & \SI{2.2}{\percent} & \SI{1.6}{\percent} & \SI{0.0}{\percent} & \SI{30.8}{\percent} & \SI{26.8}{\percent} & \SI{33.5}{\percent} & \SI{26.5}{\percent} & \SI{17.7}{\percent} & \SI{6.2}{\percent} \\
            Fev-bench & \SI{9.4}{\percent} & \SI{8.4}{\percent} & \SI{6.6}{\percent} & \SI{6.7}{\percent} & \SI{6.8}{\percent} & \SI{5.8}{\percent} & \SI{14.2}{\percent} & \SI{13.3}{\percent} & \SI{0.0}{\percent} \\
            \midrule
            Average & \SI{2.7}{\percent} & \SI{1.8}{\percent} & \SI{1.0}{\percent} & \SI{6.7}{\percent} & \SI{5.3}{\percent} & \SI{9.5}{\percent} & \SI{7.7}{\percent} & \SI{6.5}{\percent} & \SI{3.4}{\percent} \\
            \bottomrule 
        \end{tabular}
        }
    \end{table}

\newpage

\subsection{Adaptive compression of diverse time series}
\label{appendix:adaptive_compression}
\begin{figure}[H]
    \centering
    \hfill
    \begin{subfigure}[b]{4.2cm}
    \centering
        \includegraphics[width=4.2cm]{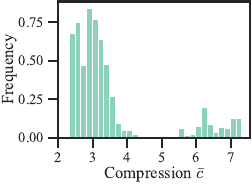}
        \captionsetup{skip=3pt}
        \caption{ETTh1}
    \end{subfigure}
    \hspace{1cm}
    \begin{subfigure}[b]{4.2cm}
        \centering
        \includegraphics[width=4.2cm]{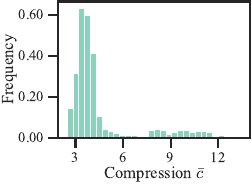}
        \captionsetup{skip=3pt}
        \caption{ETTm1}
    \end{subfigure}
    \hspace{1cm}
    \begin{subfigure}[b]{4.2cm}
        \centering
        \includegraphics[width=4.2cm]{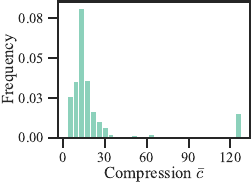}
        \captionsetup{skip=3pt}
        \caption{Weather}
    \end{subfigure}
    \hfill
    \vspace{0.2cm}
    \\
    \begin{subfigure}[b]{4.2cm}
        \centering
        \includegraphics[width=4.2cm]{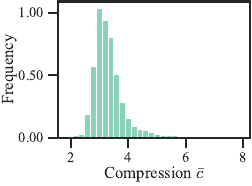}
        \captionsetup{skip=3pt}
        \caption{Electricity}
    \end{subfigure}
    \hspace{1cm}
    \begin{subfigure}[b]{4.2cm}
        \centering
        \includegraphics[width=4.2cm]{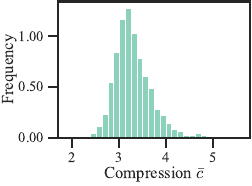}
        \captionsetup{skip=3pt}
        \caption{Traffic}
    \end{subfigure}
    \vspace{-3pt}
    \caption{Histograms showing variable compressions of our medium tokenizer within \num{5} datasets.}
    \label{fig_appendix:hist_compression}
\end{figure}

\begin{figure}[H]
    \centering
    \begin{subfigure}[b]{4.2cm}
        \centering
        \includegraphics[width=4.2cm]{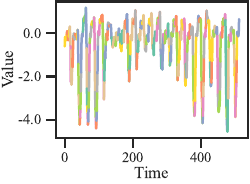}
        \captionsetup{skip=3pt}
        \caption{ETTh1; $\bar{c}=2.45$}
    \end{subfigure}
    \hspace{0.2cm}
    \begin{subfigure}[b]{4.2cm}
        \centering
        \includegraphics[width=4.2cm]{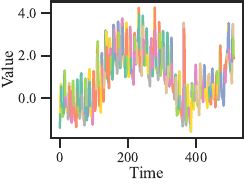}
        \captionsetup{skip=3pt}
        \caption{Electricity; $\bar{c}=2.46$}
    \end{subfigure}
    \hspace{0.2cm}
    \begin{subfigure}[b]{4.2cm}
        \centering
        \includegraphics[width=4.2cm]{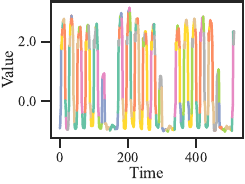}
        \captionsetup{skip=3pt}
        \caption{Electricity; $\bar{c}=3.24$}
    \end{subfigure}

    \vspace{0.3cm}
    \begin{subfigure}[b]{4.2cm}
        \centering
        \includegraphics[width=4.2cm]{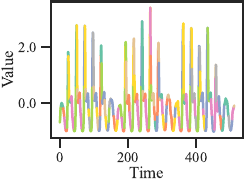}
        \captionsetup{skip=3pt}
        \caption{Traffic; $\bar{c}=3.56$}
    \end{subfigure}
    \hspace{0.2cm}
    \begin{subfigure}[b]{4.2cm}
        \centering
        \includegraphics[width=4.2cm]{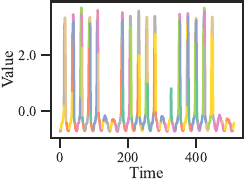}
        \captionsetup{skip=3pt}
        \caption{Traffic; $\bar{c}=3.74$}
    \end{subfigure}
    \hspace{0.2cm}
    \begin{subfigure}[b]{4.2cm}
        \centering
        \includegraphics[width=4.2cm]{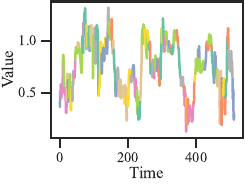}
        \captionsetup{skip=3pt}
        \caption{ETTm1; $\bar{c}=5.02$}
    \end{subfigure}

    \vspace{0.3cm}
    \begin{subfigure}[b]{4.2cm}
        \centering
        \includegraphics[width=4.2cm]{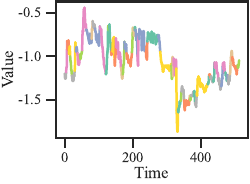}
        \captionsetup{skip=3pt}
        \caption{ETTh1; $\bar{c}=7.11$}
    \end{subfigure}
    \hspace{0.2cm}
    \begin{subfigure}[b]{4.2cm}
        \centering
        \includegraphics[width=4.2cm]{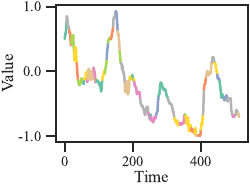}
        \captionsetup{skip=3pt}
        \caption{Weather; $\bar{c}=8.13$}
    \end{subfigure}
    \hspace{0.2cm}
    \begin{subfigure}[b]{4.2cm}
        \centering
        \includegraphics[width=4.2cm]{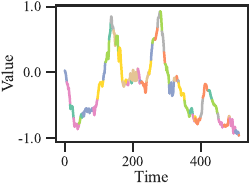}
        \captionsetup{skip=3pt}
        \caption{$\bar{c}=8.98$}
    \end{subfigure}

    \vspace{0.3cm}
    \begin{subfigure}[b]{4.2cm}
        \centering
        \includegraphics[width=4.2cm]{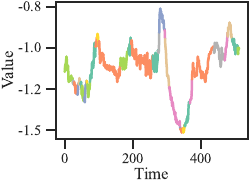}
        \captionsetup{skip=3pt}
        \caption{ETTm1; $\bar{c}=14.63$}
    \end{subfigure}
    \hspace{0.2cm}
    \begin{subfigure}[b]{4.2cm}
        \centering
        \includegraphics[width=4.2cm]{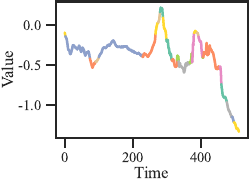}
        \captionsetup{skip=3pt}
        \caption{Weather; $\bar{c}=16.52$}
    \end{subfigure}
    \hspace{0.2cm}
    \begin{subfigure}[b]{4.2cm}
        \centering
        \includegraphics[width=4.2cm]{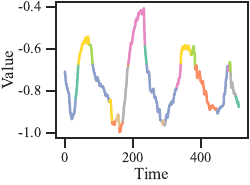}
        \captionsetup{skip=3pt}
        \caption{Weather; $\bar{c}=18.95$}
    \end{subfigure}
    
    \vspace{0.3cm}
    \begin{subfigure}[b]{4.2cm}
        \centering
        \includegraphics[width=4.2cm]{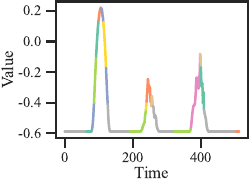}
        \captionsetup{skip=3pt}
        \caption{Weather; $\bar{c}=22.26$}
    \end{subfigure}
    \hspace{0.2cm}
    \begin{subfigure}[b]{4.2cm}
        \centering
        \includegraphics[width=4.2cm]{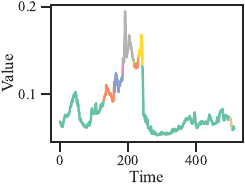}
        \captionsetup{skip=3pt}
        \caption{Weather; $\bar{c}=42.67$}
    \end{subfigure}
    \hspace{0.2cm}
    \begin{subfigure}[b]{4.2cm}
        \centering
        \includegraphics[width=4.2cm]{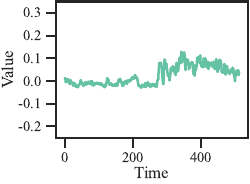}
        \captionsetup{skip=3pt}
        \caption{Weather; $\bar{c}=128.00$}
    \end{subfigure}
    \vspace{-3pt}
    \caption{Our medium tokenizer exploits periodically recurring motifs and compresses time series adaptively depending on pattern complexity on \num{5} datasets. Specifically, \textbf{(o)} highlights the noise rejection ability of discretization.}
    \label{fig_appendix:tokenized_timeseries}
\end{figure}

Here, we present the complete results of our investigations on adaptive compression of our motif-based tokenization approach. In \cref{fig_appendix:hist_compression}, we explore variable compression within the same dataset and show tokenized time series for visual inspection in \cref{fig_appendix:tokenized_timeseries}. We further investigate relations between input and generation compression. Finally, we list compression rates of patch-based literature models as a reference.

\begin{figure}[H]
    \centering
    \hfill
    \begin{subfigure}[b]{4.2cm}
    \centering
        \includegraphics[width=4.2cm]{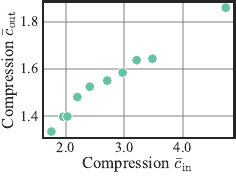}
        \captionsetup{skip=3pt}
        \caption{ETTh1}
    \end{subfigure}
    \hspace{1cm}
    \begin{subfigure}[b]{4.2cm}
        \centering
        \includegraphics[width=4.2cm]{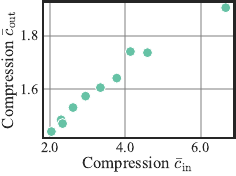}
        \captionsetup{skip=3pt}
        \caption{ETTm1}
    \end{subfigure}
    \hspace{1cm}
    \begin{subfigure}[b]{4.2cm}
        \centering
        \includegraphics[width=4.2cm]{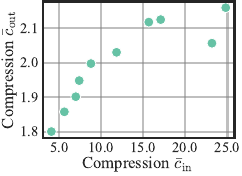}
        \captionsetup{skip=3pt}
        \caption{Weather}
    \end{subfigure}
    \hfill
    \vspace{0.2cm}
    \\
    \begin{subfigure}[b]{4.2cm}
        \centering
        \includegraphics[width=4.2cm]{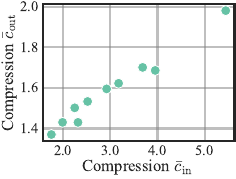}
        \captionsetup{skip=3pt}
        \caption{Electricity}
    \end{subfigure}
    \hspace{1cm}
    \begin{subfigure}[b]{4.2cm}
        \centering
        \includegraphics[width=4.2cm]{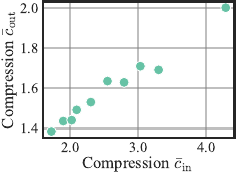}
        \captionsetup{skip=3pt}
        \caption{Traffic}
    \end{subfigure}
    \vspace{-3pt}
    \caption{Comparison of input and generation compression of small models and multiple tokenizers on \num{5} datasets. Please note that efficiency gains in \cref{tab:tok_chronos,tab_appendix:main_results,fig:tok_electricity,fig_appendix:main_results} relate to the more conservative generation compression.}
    \label{fig_appendix:input_generation_compression}
\end{figure}

\textbf{Input and generation compression}\hspace{1.8mm} We conduct additional experiments to explore relations between input compression and the model's generations. The model can either predict long motifs with high compression directly or sequences of their shorter components during autoregressive generation, as we describe in \cref{sec:explainability_patterns}. Therefore, we expect a greater input compression $\bar{c}_{\mathrm{in}}$ compared to generation compression $\bar{c}_{\mathrm{out}}$. We utilize different tokenizers in small models for this experiment. In line with our hypothesis, we find correlations between input and generation compression on all \num{5} datasets in \cref{fig_appendix:input_generation_compression}. More complex input tokens generally benefit the prediction of longer motifs.

\newpage

\textbf{Compression rates of patch-based models}\hspace{1.8mm} In \cref{tab:patch_literature_compression_rates} we list compression rates of patch-based models in recent literature, resulting from different patch length and stride combinations. For some works that experiment with multiple patch lengths, we show the authors' preferred values.

   \begin{table}[H]
      \caption{Compression rates of patch-based literature models.}
      \label{tab:patch_literature_compression_rates}
      \centering
      \vspace{-0.5\baselineskip}
        \begin{tabular}{lr}
            \toprule 
            Architecture & Compression $\bar{c}$\\ 
            \midrule 
            SDformer \citep{chen2024sdformer} & \num{2}, \num{4}\\
            TOTEM \citep{talukder2024totem} & \num{4}\\
            MOMENT \citep{goswami2024moment} & \num{8}\\
            PatchTST \citep{nie2023patchtst} & \num{8}\\
            TimeXer \citep{wang2024timexer} & \num{16}\\
            UniTS \citep{gao2024units} & \num{16}\\
            Sundial \citep{liu2025sundial} & \num{16}\\
            Moirai-MoE \citep{liu2024moiraimoe} & \num{16}\\
            \bottomrule 
        \end{tabular}
    \end{table}

\newpage

\subsection{Vocabulary complexity and generalization}
\label{appendix:pattern_complexity}
In \cref{appendix_fig:pmin_ablation,tab_appendix:pmin_tokenizers}, we provide full results of our investigations on token occurrence. We offer additional insights into the hierarchy of motifs and the vocabulary generation process.

\vspace{0.2cm}
\begin{figure}[htbp]
        \hfill
        \begin{subfigure}[b]{4.2cm}
        \centering
        \includegraphics[width=4.2cm]{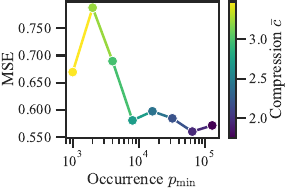}
        \captionsetup{skip=3pt}
        \caption{ETTh1}
    \end{subfigure}
    \hspace{1cm}
    \begin{subfigure}[b]{4.2cm}
        \centering
        \includegraphics[width=4.2cm]{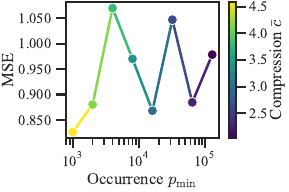}
        \captionsetup{skip=3pt}
        \caption{ETTm1}
    \end{subfigure}
    \hspace{1cm}
    \centering
        \begin{subfigure}[b]{4.2cm}
        \centering
        \includegraphics[width=4.2cm]{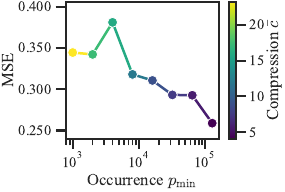}
        \captionsetup{skip=3pt}
        \caption{Weather}
    \end{subfigure}
    \hfill
    \vspace{0.3cm}
    \\
    \begin{subfigure}[b]{4.2cm}
        \centering
        \includegraphics[width=4.2cm]{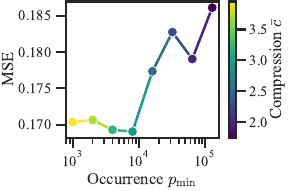}
        \captionsetup{skip=3pt}
        \caption{Electricity}
    \end{subfigure}
    \hspace{1cm}
    \begin{subfigure}[b]{4.2cm}
        \centering
        \includegraphics[width=4.2cm]{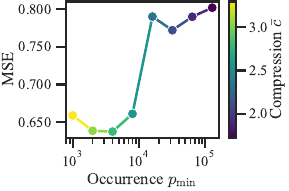}
        \captionsetup{skip=3pt}
        \caption{Traffic}
    \end{subfigure}
    \vspace{-3pt}
    \caption{Varying token occurrence $p_{\mathrm{min}}$ influences forecasting quality for small models on \num{5} datasets.}
    \label{appendix_fig:pmin_ablation}
\end{figure}

   \begin{table}[H]
      \caption{Tokenizers on the Chronos dataset with different token occurrence, vocabulary size, and compression.}
      \vspace{-0.5\baselineskip}
      \label{tab_appendix:pmin_tokenizers}
      \centering
        \begin{tabular}{lrr}
            \toprule 
            $p_{\mathrm{min}}$ & \multicolumn{1}{c}{$|\mathcal{V}|$} & \multicolumn{1}{c}{$\bar{c}$}\\ 
            \midrule 
            \num{1000} & \num{1675} & \num{3.18}\\
            \num{2000} & \num{993} & \num{2.95}\\
            \num{4000} & \num{604} & \num{2.73}\\
            \num{8000} & \num{373} & \num{2.50}\\
            \num{16000} & \num{237} & \num{2.29}\\
            \num{32000} & \num{158} & \num{2.08}\\
            \num{64000} & \num{108} & \num{1.86}\\
            \num{128000} & \num{78} & \num{1.66}\\

            \bottomrule 
        \end{tabular}
    \end{table}

\begin{figure}[h]
    \centering
    \includegraphics[width=2.37in,trim={0in 0in 0in 0in},clip]{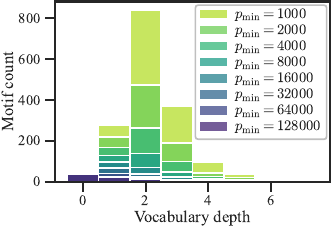}
    \vspace{-3pt}
    \caption{Motif hierarchy for vocabularies of different complexity.}
    \label{fig_appendix:vocab_hierarchy}
\end{figure}

\textbf{Hierarchy of motifs}\hspace{1.8mm}  Motif-based tokenization utilizes a vocabulary of hierarchical patterns. Here, we explore the hierarchy of motifs evolving with more complex vocabularies. To this end, we vary the minimum occurrence threshold $p_{\mathrm{min}}$. 
Naturally, a lower occurrence threshold results in larger vocabularies. These vocabularies exhibit more complex patterns generated by a greater number of recursive merges in \cref{fig_appendix:vocab_hierarchy}. Due to its hierarchical structure, motif length grows exponentially with vocabulary depth, enabling substantial compression.

\newpage

\begin{figure}[h]
    \centering
    \includegraphics[width=5.5in,trim={0in 0in 0in 0in},clip]{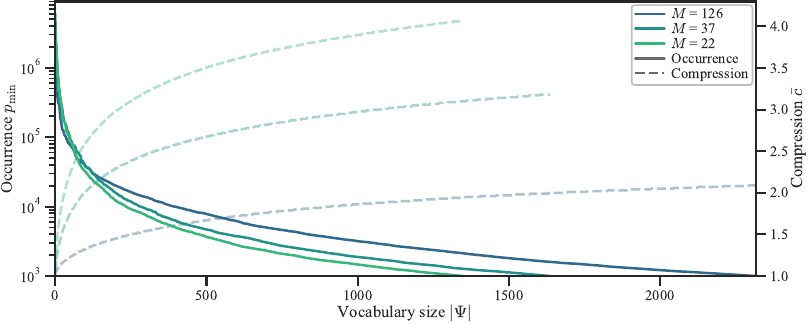}
    \vspace{-3pt}
    \caption{Influence of quantization bins $M$ and token occurrence $p_{\mathrm{min}}$ on vocabulary size $|\Psi|$ and compression $\bar{c}$ for tokenizers on the Chronos dataset.}
    \label{fig_appendix:vocab_generation}
\end{figure}

\textbf{Vocabulary generation process}\hspace{1.8mm}  Here, we further highlight the influence of quantization granularity and token occurrence on compression and vocabulary complexity. In \cref{fig_appendix:vocab_generation}, we show the iterative process of finding longer motifs with higher compression $\bar{c}$ during vocabulary generation of $\Psi$. These more complex motifs, however, are more specialized and occur less often ($p_{\mathrm{min}}$). A lower number of quantization bins $M$ results in smaller, less data-specific vocabularies with higher compression. 

\vspace{5pt}

\newpage

\subsection{Robustness to noise, non-stationary time series, and transients}
\label{appendix:noise_robustness}
Robustness to noise is of high relevance when processing real-world time series. Further, the changing distribution of non-stationary time series or extreme values might hinder effective motif-based tokenization. Here, we explore our tokenizer's generalization to noise, distribution shifts, and transients in more detail.

\begin{figure}[H]
    \centering
    \hfill
    \begin{subfigure}[b]{4.2cm}
    \centering
        \includegraphics[width=4.2cm]{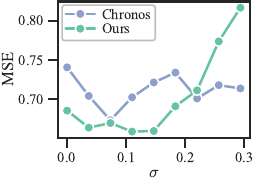}
        \captionsetup{skip=3pt}
        \caption{ETTh1}
    \end{subfigure}
    \hspace{1cm}
    \begin{subfigure}[b]{4.2cm}
        \centering
        \includegraphics[width=4.2cm]{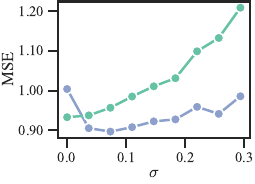}
        \captionsetup{skip=3pt}
        \caption{ETTm1}
    \end{subfigure}
    \hspace{1cm}
    \begin{subfigure}[b]{4.2cm}
        \centering
        \includegraphics[width=4.2cm]{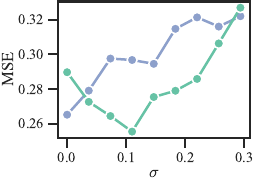}
        \captionsetup{skip=3pt}
        \caption{Weather}
    \end{subfigure}
    \hfill
    \vspace{0.3cm}
    \\
    \begin{subfigure}[b]{4.2cm}
        \centering
        \includegraphics[width=4.2cm]{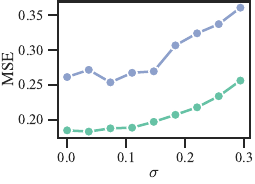}
        \captionsetup{skip=3pt}
        \caption{Electricity}
    \end{subfigure}
    \hspace{1cm}
    \begin{subfigure}[b]{4.2cm}
        \centering
        \includegraphics[width=4.2cm]{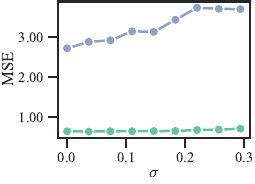}
        \captionsetup{skip=3pt}
        \caption{Traffic}
    \end{subfigure}
    \vspace{-3pt}
    \caption{Resistance of Chronos models and our high compression tokenizer to  Gaussian noise with standard deviation $\sigma$ on \num{5} datasets.}
    \label{fig_appendix:noise_rejection}
\end{figure}

\textbf{Robustness to noise}\hspace{1.8mm} To explore the noise rejection capability of motif-based tokenization, we injected different levels of Gaussian noise into the raw input sequence before tokenization. We compare our high compression tokenizer in a small model to the respective Chronos baseline. \\
Our results in \cref{fig_appendix:noise_rejection} show that our motif-based tokenization substantially outperforms Chronos models on noisy data. Further, its noise resistance is more predictable. We argue that our method is more robust to noise due to its coarser quantization granularity. At the same time, our more expressive motif representation mitigates the larger discretization error.
Note that adding noise with up to $\sigma=0.3$ to our normalized input data with unit standard deviation is a severe disturbance.

\begin{figure}[H]
    \centering
    
    \begin{subfigure}[b]{4.2cm}
    \centering
        \includegraphics[width=4.2cm]{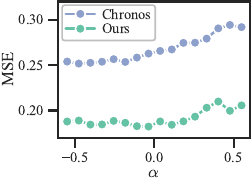}
        \captionsetup{skip=3pt}
        \caption{Linear trend in range $[-\alpha, \alpha]$}
    \end{subfigure}
    \hspace{1.0cm}
    \begin{subfigure}[b]{4.2cm}
        \centering
        \includegraphics[width=4.2cm]{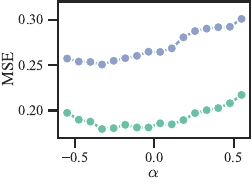}
        \captionsetup{skip=3pt}
        \caption{\makebox[0.6\textwidth][l]{Exponential trend in range $[0, \alpha]$}}
    \end{subfigure}
    \vspace{-3pt}
    \caption{Generalization of Chronos models and our high compression tokenizer to non-stationary time series with linear and exponential trends on the Electricity dataset.}
    \label{fig_appendix:trend_rejection}
\end{figure}

\textbf{Generalization to non-stationary data}\hspace{1.8mm} In practice, trends on long non-stationary time series might hinder effective motif encoding. To explore this, we introduce linear and exponential trends into our evaluation datasets. We utilize our high compression tokenizer in a small model and the corresponding Chronos baseline for this experiment. \\
In \cref{fig_appendix:trend_rejection}, our method shows a similar robustness to non-stationary data compared to the Chronos baseline, even for large trends. We conclude that our motif-based tokenization is well applicable to non-stationary time series and long sequences. Note that applying trends with up to $|\alpha|=0.5$ to our normalized data is a large disturbance.

\vspace{-0.25\baselineskip}

\begin{figure}[H]
    \centering
    \includegraphics[width=2.475in,trim={0in 0in 0in 0in},clip]{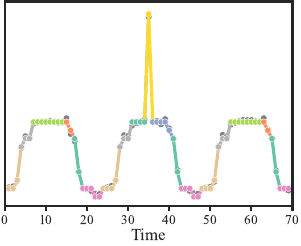}
    \vspace{-3pt}
    \caption{Introducing a transient (yellow) in the center period leads to local changes in tokenization of time series samples (gray) compared to the other periods on the Electricity dataset.}
    \label{fig_appendix:transient_visualized}
\end{figure}

\newpage
\textbf{Robustness to transients}\hspace{1.8mm} Extreme values might occur in real-world time series. Here, we explore the robustness of our motif-based tokenization to outliers. To this end, we augment the input time series such that every sample has a \SI{1}{\percent} probability to be a positive or negative transient with amplitude \num{3}. For our normalized time series with unit standard deviation, this is a severe disturbance. We analyze our tokenizer with medium compression and models in size small on the Electricity dataset.\\
While Chronos models cannot effectively handle extreme values, our motif-based tokenization is substantially more robust to outliers as our results in \cref{tab_appendix:transients} show. The MSE of our models increases by only \SI{19.4}{\percent}, compared to \SI{52.5}{\percent} for Chronos models. When encountering unknown patterns that are not in our tokenizer’s motif vocabulary, such as transients, our tokenizer falls back to sample-based tokenization, as illustrated in \cref{fig_appendix:transient_visualized}. This fallback ensures that motif-based tokenization cannot overlook individual samples by design. Consequently, compression is slightly reduced by \SI{14.4}{\percent} when outliers are introduced. \\
Note that within the tokenization range, the same maximum quantization error applies regardless of whether a sample is common or an outlier, as described in \cref{sec_method:quantization}.

\begin{table}[H]
       \small
      \caption{MSE and compression $\bar{c}$ for Chronos and our medium compression tokenizer on the Electricity dataset with and without transient augmentation.}
      \vspace{-0.5\baselineskip}
      \label{tab_appendix:transients}
      \centering
        \begin{tabular}{lccc}
            \toprule
            \multirow{2}{*}{Augmentation} &  \multicolumn{1}{c}{Chronos} &\multicolumn{2}{c}{Ours}  \\
            \cmidrule{2-2}\cmidrule{3-4}
            & MSE & MSE & $\bar{c}$\\
            \midrule 
            Without transients & \num{0.261} & \num{0.170} & \num{3.95}\\
            With transients & \num{0.398} & \num{0.203} & \num{3.38}\\
            \bottomrule 
        \end{tabular}
    \end{table}

\newpage

\subsection{Training dataset size}
\label{appendix:training_set_size}

We vary the dataset size for building our vocabulary of motifs $\Psi$ and provide full results here. To this end, we utilize our three tokenizers with low, medium, and high compression and report vocabulary statistics in \cref{tab_appendix:training_set_tok} and forecasting quality in \cref{tab_appendix:training_set_mse}. For conditional decoding, we observe similar behavior when estimating conditional distributions $\hat{\Omega}$ in \cref{tab_appendix:training_set_cond_dec}. Here, a larger subset also leads to best representations.

\begin{table}[H]
      \caption{Vocabulary statistics for \num{3} tokenizers trained on Chronos subsets of varying sizes $N$.}
      \vspace{-0.5\baselineskip}
      \label{tab_appendix:training_set_tok}
      \centering
        \begin{tabular}{lrlrlrlrl}
            \toprule 
            \multirow{2}{*}{Compression} &  \multicolumn{2}{c}{$N =$ 1\,k} & \multicolumn{2}{c}{$N =$ 10\,k} & \multicolumn{2}{c}{$N =$ 100\,k} & \multicolumn{2}{c}{$N =$ 1\,M} \\
            \cmidrule{2-3}\cmidrule{4-5}\cmidrule{6-7}\cmidrule{8-9}
            & \multicolumn{1}{c}{$|\mathcal{V}|$} & \multicolumn{1}{c}{$\bar{c}$} & \multicolumn{1}{c}{$|\mathcal{V}|$} & \multicolumn{1}{c}{$\bar{c}$} &\multicolumn{1}{c}{$|\mathcal{V}|$} & \multicolumn{1}{c}{$\bar{c}$} &\multicolumn{1}{c}{$|\mathcal{V}|$} & \multicolumn{1}{c}{$\bar{c}$}\\ 
            \midrule 
            low & \num{2789}& \num{2.11}& \num{2461}& \num{2.08}& \num{2445}& \num{2.08}& \num{2441}& \num{2.09}\\
            medium &  \num{1974}& \num{3.24}& \num{1707}& \num{3.16}& \num{1675}& \num{3.18}& \num{1681}& \num{3.18}\\
            high &  \num{1618}& \num{4.14}& \num{1392}& \num{4.05}& \num{1373}& \num{4.06}& \num{1360}& \num{4.06}\\
            \bottomrule 
        \end{tabular}
    \end{table}

\begin{table}[H] 
      \caption{Forecasting quality (MSE) on \num{5} evaluation datasets for \num{3} tokenizers trained on Chronos subsets of varying sizes $N$.}
      \vspace{-0.5\baselineskip}
      \label{tab_appendix:training_set_mse}
      \centering
        \begin{tabular}{llrrrr}
            \toprule 
            Dataset & Compression & $N =$ 1\,k & $N =$ 10\,k & $N =$ 100\,k & $N =$ 1\,M\\
            \midrule 
            \multirow{3}{*}{ETTh1} & low & \num{0.712} & \num{0.712} & \num{0.656} & \num{0.659}\\
            & medium & \num{0.562} & \num{0.698} & \num{0.669} & \num{0.615}\\
            & high & \num{0.751} & \num{0.712} & \num{0.686} & \num{0.593}\\
            \midrule 
            \multirow{3}{*}{ETTm1} & low & \num{0.944} & \num{0.919} & \num{0.934} & \num{0.913}\\
            & medium & \num{0.877} & \num{0.898} & \num{0.826} & \num{0.857}\\
            & high & \num{0.985} & \num{0.819} & \num{0.933} & \num{0.821}\\
            \midrule 
            \multirow{3}{*}{Weather} & low & \num{0.473} & \num{0.538} & \num{0.463} & \num{0.474}\\
            & medium & \num{0.342} & \num{0.310} & \num{0.344} & \num{0.333}\\
            & high & \num{0.355} & \num{0.333} & \num{0.290} & \num{0.307}\\
            \midrule 
            \multirow{3}{*}{Electricity} & low & \num{0.178} & \num{0.183} & \num{0.176} & \num{0.175}\\
            & medium & \num{0.173} & \num{0.167} & \num{0.170} & \num{0.164}\\
            & high & \num{0.191} & \num{0.176} & \num{0.185} & \num{0.178}\\
            \midrule 
            \multirow{3}{*}{Traffic} & low & \num{0.724} & \num{0.680} & \num{0.693} & \num{0.671}\\
            & medium & \num{0.643} & \num{0.639} & \num{0.659} & \num{0.622}\\
            & high & \num{0.625} & \num{0.621} & \num{0.646} & \num{0.620}\\
            \bottomrule 
        \end{tabular}
    \end{table}

\begin{table}[H]
      \caption{Influence of dataset size $N$ on estimating conditional decoding distributions $\hat{\Omega}$ for small models on the Electricity dataset.}
      \vspace{-0.5\baselineskip}
      \label{tab_appendix:training_set_cond_dec}
      \centering
        \begin{tabular}{lr}
            \toprule 
            $N$ &  MSE\\
            \midrule 
             Normal decoding & \num{0.170}\\
             1\,k & \num{0.168}\\
             10\,k & \num{0.162}\\
             100\,k & \num{0.161}\\
            \bottomrule 
        \end{tabular}
    \end{table}

\newpage

\subsection{Learned token representations}
Here, we present our comprehensive explainability analysis of the learned motif representations.

\vspace{0.3cm}
\begin{figure}[htbp]
    \centering
    \begin{subfigure}[b]{4.2cm}
        \centering
        \includegraphics[width=4.2cm]{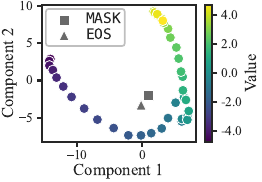}
        \captionsetup{skip=3pt}
        \caption{Symbol values}
    \end{subfigure}
    \hspace{0.2cm}
    \begin{subfigure}[b]{4.2cm}
        \centering
        \includegraphics[width=4.2cm]{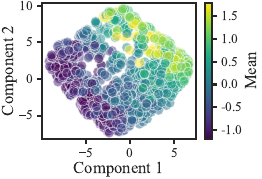}
        \captionsetup{skip=3pt}
        \caption{Motif means}
    \end{subfigure}
    \hspace{0.2cm}
    \begin{subfigure}[b]{4.2cm}
        \centering
        \includegraphics[width=4.2cm]{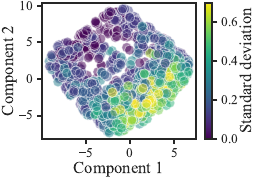}
        \captionsetup{skip=3pt}
        \caption{Motif standard deviations}
    \end{subfigure}

        \vspace{0.3cm}
    \begin{subfigure}[b]{4.2cm}
        \centering
        \includegraphics[width=4.2cm]{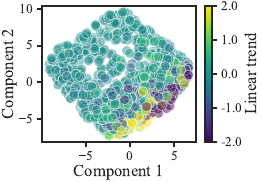}
        \captionsetup{skip=3pt}
        \caption{Linear trend of motifs}
    \end{subfigure}
    \hspace{0.2cm}
    \begin{subfigure}[b]{4.2cm}
        \centering
        \includegraphics[width=4.2cm]{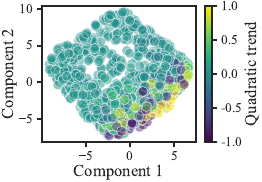}
        \captionsetup{skip=3pt}
        \caption{Quadratic trend of motifs}
    \end{subfigure}
        \hspace{0.2cm}
    \begin{subfigure}[b]{4.2cm}
        \centering
        \includegraphics[width=4.2cm]{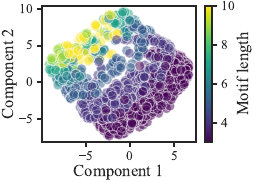}
        \captionsetup{skip=3pt}
        \caption{Motif lengths}
    \end{subfigure}
    \vspace{-3pt}
    \caption{Principal component analysis of token embeddings of our medium tokenizer in a small model.}
    \label{fig_appendix:pca_e}
\end{figure}

\vspace{1cm}
\begin{figure}[h]
    \centering
    \includegraphics[width=2.8in,trim={0in 0in 0in 0in},clip]{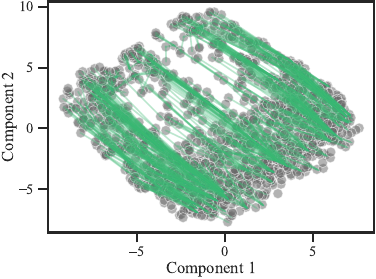}
    \vspace{-3pt}
    \caption{Parent-child token relations analyzed through principal component analysis of token embeddings from our medium tokenizer in a small model. Children and their first parents are connected.}
    \label{fig_appendix:pca_parent_child}
\end{figure}

\newpage

\subsection{Learned motifs}
\label{appendix:motifs_visualized}
Our tokenizer employs a vocabulary of frequent motifs to encode time series. To enhance interpretability, we illustrate selected patterns learned by our tokenizer with medium compression (see \cref{tab:tokenizers}). Note that the vocabulary also includes shifted and scaled variants of these motifs along the y-axis.

\vspace{0.3cm}
\begin{figure}[H]
    \centering
    \includegraphics[width=\textwidth,trim={0in 0in 0in 0in},clip]{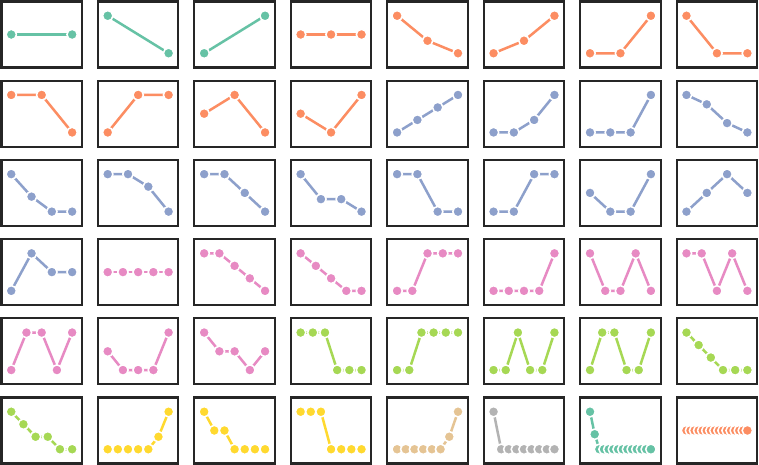}
    \vspace{-3pt}
    \caption{Visualization of motifs which our medium compression tokenizer uses to encode time series. Colors indicate motif length.}
    \label{fig_appendix:motifs_visualized}
\end{figure}

\end{document}